\documentclass[10pt,twocolumn,letterpaper]{article}

\usepackage{cvpr}
\usepackage{titling}
\usepackage{times}
\usepackage{epsfig}
\usepackage{graphicx}
\usepackage{amsmath}
\usepackage{amssymb}
\usepackage{amsthm}
\usepackage{algorithm}
\usepackage{caption}
\usepackage{subcaption}
\usepackage{enumitem}
\usepackage[noend]{algpseudocode}

\makeatletter
\def\BState{\State\hskip-\ALG@thistlm}
\newcommand*\bigcdot{\mathpalette\bigcdot@{1.0}}
\newcommand*\bigcdot@[2]{\mathbin{\vcenter{\hbox{\scalebox{#2}{$\m@th#1\bullet$}}}}}
\makeatother

\usepackage[breaklinks=true,letterpaper=true,colorlinks,bookmarks=false]{hyperref}

 \cvprfinalcopy 

\def\cvprPaperID{1610} 

\ifcvprfinal\fi
\begin{document}

\title{C2AE: Class Conditioned Auto-Encoder for Open-set Recognition}

\author{Poojan Oza and Vishal M. Patel\\\
Department of Electrical and Computer Engineering\\
Johns Hopkins University, 3400 N. Charles St, Baltimore, MD 21218, USA\\
{\tt\small poza2@jhu.edu, vpatel36@jhu.edu}
}
\date{}
\maketitle

\begin{abstract}

Models trained for classification often assume that all testing classes are known while training. As a result, when presented with an unknown class during testing, such \textit{closed-set} assumption forces the model to classify it as one of the known classes. However, in a real world scenario, classification models are likely to encounter such examples. Hence, identifying those examples as unknown becomes critical to model performance. A potential solution to overcome this problem lies in a class of learning problems known as open-set recognition. It refers to the problem of identifying the unknown classes during testing, while maintaining performance on the known classes. In this paper, we propose an open-set recognition algorithm using class conditioned auto-encoders with novel training and testing methodology. In contrast to previous methods, training procedure is divided in two sub-tasks, 1. closed-set classification and, 2. open-set identification (\textit{i.e.} identifying a class as known or unknown). Encoder learns the first task following the closed-set classification training pipeline, whereas decoder learns the second task by reconstructing conditioned on class identity. Furthermore, we model reconstruction errors using the Extreme Value Theory of statistical modeling to find the threshold for identifying known/unknown class samples. Experiments performed on multiple image classification datasets show proposed method performs significantly better than state of the art.

\end{abstract}

\section{Introduction}
\label{sec:intro}

\begin{figure}[t]
\begin{center}
   \includegraphics[width=0.8\linewidth]{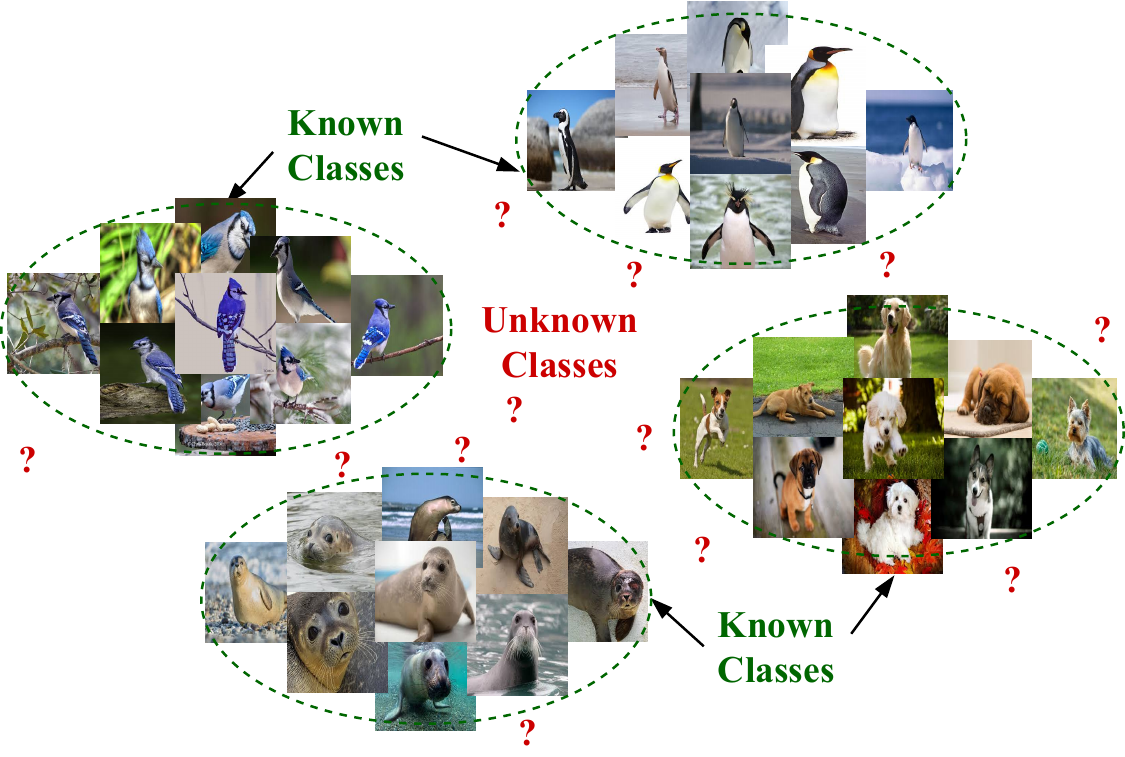}
\end{center}
   \vskip -15.0pt \caption{Open-set recognition problem: Data samples from Blue Jay, Seal, Dog and Penguin are from the known class set ($\mathcal{K}$). Also, many classes not known during training, will be present at testing, \textit{i.e.}, samples from unknown class set ($\mathcal{U}$). The goal is to correctly classify any sample coming from set $\mathcal{K}$, as either Blue Jay, Seal, Dog or Penguin and identify samples coming from $\mathcal{U}$ as unknown.}
\label{fig:open_set}
\end{figure}

Recent advancements in computer vision have resulted in significant improvements for image classification systems \cite{resnet16}, \cite{alexnet12}, \cite{senet17}, \cite{simonyan2014very}. Especially the rise of Deep Convolutional Neural Network has resulted in classification error rates surpassing the human-level performance \cite{he2015delving}. These promising results, enable their potential use in many real world applications. However, when deployed in a real world scenario, such systems are likely to observe samples from classes not seen during training (\textit{i.e.} unknown classes also referred as ``\textit{unknown unknowns}'' \cite{scheirer2013toward}). Since, the traditional training methods follow this closed-set assumption, the classification systems observing any unknown class samples are forced to recognize it as one of the known classes. As a result, it affects the performance of these systems, as evidenced by Jain \emph{et al.} with digit recognition example. Hence, it becomes critical to correctly identify test samples as either known or unknown for a classification model. This problem setting of identifying test samples as known/unknown and simultaneously correctly classifying all of known classes, is referred to as open-set recognition \cite{scheirer2013toward}. Fig.~\ref{fig:open_set} illustrates a typical example of classification in the open-set problem setting.

In an open-set problem setting, it becomes challenging to identify unknown samples due to the incomplete knowledge of the world during training (\textit{i.e.} only the known classes are accessible).  To overcome this problem many open-set methods in the literature \cite{bendale2016towards}, \cite{scheirer2014probability}, \cite{zhang2017sparse}, \cite{shu2017doc} adopt recognition score based thresholding models. However, when using these models one needs to deal with two key questions, \textit{ 1) what is a good score for open-set identification?} (\textit{i.e.}, identifying a class as known or unknown), and given a score, \textit{ 2) what is a good operating threshold for the model?}. There have been many methods that explore these questions in the context of traditional methods such as Support Vector Machines \cite{scheirer2013toward}, \cite{scheirer2014probability}, Nearest Neighbors \cite{junior2017nearest}, \cite{bendale2015towards} and Sparse Representation  \cite{zhang2017sparse}. However, these questions are relatively unexplored in the context of deep neural networks \cite{shu2017doc}, \cite{bendale2016towards}, \cite{neal2018open}, \cite{ge2017generative}, \cite{dhamija2018reducing}.


Even-though deep neural networks are powerful in learning highly discriminative representations, they still suffer from performance degradation in the open-set setting \cite{bendale2016towards}. In a naive approach, one could apply a thresholding model on SoftMax scores. However, as shown by experiments in \cite{bendale2016towards}, that model is sub-optimal for open-set identification. A few methods have been proposed to better adapt the SoftMax scores for open-set setting. Bendale \emph{et al.} proposed a calibration strategy to update SoftMax scores using extreme value modeling \cite{bendale2016towards}. Other strategies, Ge \emph{et al.} \cite{ge2017generative} and Lawrence \emph{et al.} \cite{neal2018open} follow data augmentation technique using Generative Adversarial Networks (GANs) \cite{goodfellow2014generative}. GANs are used to synthesize open-set samples and later used to fine-tuning to adapt SoftMax/OpenMax scores for open-set setting. Shu \emph{et al.} \cite{shu2017doc} introduced a novel sigmoid-based loss function for training the neural network to get better scores for open-set identification. 

All of these methods modify the SoftMax scores, so that it can perform both open-set identification and maintain its classification accuracy. However, it is extremely challenging to find a single such score measure, that can perform both. In Contrast to these methods, in proposed approach the training procedure for open-set recognition using class conditional auto-encoders, is divided it into two sub-tasks, 1. closed-set classification, and 2. open-set identification. These sub-tasks are trained separately in a stage-wise manner.  Experiments show that such approach provides good open-set identification scores and it is possible to find a good operating threshold using the proposed training and testing strategy. 

In summary, this paper makes following contributions,
\begin{itemize}[topsep=0pt,noitemsep,leftmargin=*]
\item A novel method for open-set recognition is proposed with novel training and testing algorithm based on class conditioned auto-encoders.
\item We show that dividing open-set problem in sub-tasks can help learn better open-set identification scores.
\item Extensive experiments are conducted on various image classification  datasets  and  comparisons are  performed  against  several  recent  state-of-the-art approaches. Furthermore, we analyze the effectiveness of the proposed method through ablation experiments.
\end{itemize}


\section{Related Work}
\label{sec:related_work}

\begin{figure*}[t!]
	\begin{center}
		\includegraphics[width=0.8\linewidth]{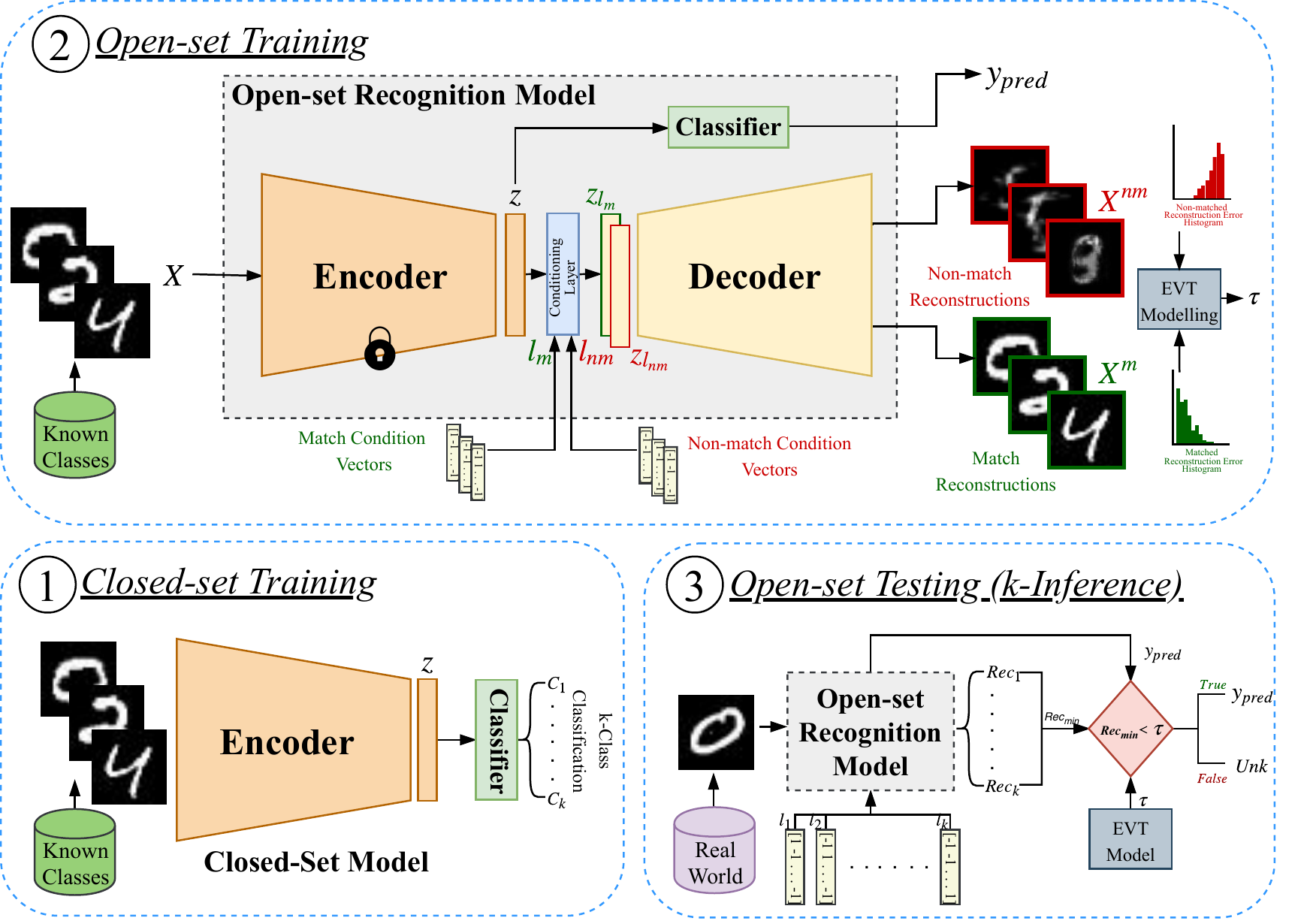}
	\end{center}
	\vskip -15.0pt \caption{Block diagram of the proposed method: \textbf{1) Closed-set training}, Encoder ($\mathcal{F}$) and Classifier ($\mathcal{C}$) are trained with the traditional classification loss. \textbf{2) Open-set Training}, To train an open-set identification model, auto-encoder network Encoder ($\mathcal{F}$) with frozen weights, and Decoder ($\mathcal{G}$), are trained to perfectly or poorly reconstruct the images depending on the label condition vector. Reconstruction errors are then modeled using the extreme value distribution to find the operating threshold of the method. \textbf{3) Open-set Testing}, Open-set recognition model produces the classification prediction ($y_{pred}$) and $k$ reconstruction errors, conditioned with each condition vector. If the minimum reconstruction error is below the threshold value obtained from the EVT model, the test sample is classified as one of the $k$ classes, or else it is classified as unknown.}
	\label{fig:block_dia}
\end{figure*}

\noindent \textbf{Open-set Recognition.} The open-set recognition methods can be broadly classified in to two categories, traditional methods and neural network-based methods. Traditional methods are based on classification models such as Support Vector Machines (SVMs), Nearest Neighbors, Sparse Representation etc. Scheirer \emph{et al.} \cite{scheirer2014probability} extended the SVM for open-set recognition by calibrating the decision scores using the extreme value distribution. Specifically, Scheirer \textit{et al.} \cite{scheirer2014probability} utilized two SVM models, one for identifying a sample as unknown (referred as CAP models) and other for traditional closed-set classification. PRM Junior \emph{et al.} \cite{junior2016specialized} proposed a nearest neighbor-based open-set recognition model utilizing the neighbor similarity as a score for open-set identification. PRM Junior \emph{et al.} later also presented specialized SVM by constraining the bias term to be negative. This strategy in the case of Radial Basis Function kernel, yields an open-set recognition model. Zhang \emph{et al.} \cite{zhang2017sparse} proposed an extension of the Sparse Representation-based Classification (SRC) algorithm for open-set recognition. Specifically, they model residuals from SRC using the Generalized-Pareto extreme value distribution to get score for open-set identification.

In neural network-based methods, one of the earliest works by Bendale \emph{et al.} \cite{bendale2016towards} introduced an open-set recognition model based on ``activation vectors" (\textit{i.e.} penultimate layer of the network). Bendale \emph{et al.} utilized meta-recognition for multi-class classification by modeling the distance from ``mean activation vector'' using the extreme value distribution. SoftMax scores are calibrated using these models for each class. These updated scores, termed as OpenMax, are then used for open-set identification. Ge \emph{et al.} \cite{ge2017generative} introduced a data augmentation approach called G-OpenMax. They generate unknown samples from the known class training data using GANs and use it to fine-tune the closed-set classification model. This helps in improving the performance for both SoftMax and OpenMax based deep network. Along the similar motivation, Neal \emph{et al.} \cite{neal2018open} proposed a data augmentation strategy called \textit{counterfacutal image generation}. This strategy also utilizes GANs to generate images that resemble known class images but belong to unknown classes. In another approach, Shu \emph{et al.} \cite{shu2017doc} proposed a $k$-sigmoid activation-based novel loss function to train the neural network. Additionally, they perform score analysis on the final layer activations to find an operating threshold, which is helpful for open-set identification. There are some variation of open-set recognition by relaxing its formulation in the form of anomaly detection \cite{oza2019active}, \cite{oza2019one}, \cite{perera2018learning} and novelty detection \cite{perera2019deep}, \cite{perera2019ocgan} etc, but for this paper we only focus on the general open-set recognition problem.

\noindent \textbf{Extreme Value Theory.} Extreme value modeling is a branch of statistics that deals with modeling of statistical extremes. The use of extreme value theory in vision tasks largely deals with post recognition score analysis \cite{perera2017extreme}, \cite{scheirer2014probability}. Often for a given recognition model the threshold to reject/accept lies in the overlap region of extremes of match and non-match score distributions \cite{shi2008modeling}. In such cases, it makes sense to model the tail of the match and non-match recognition scores as one of the extreme value distributions. Hence, many visual recognition methods including some described above, utilize extreme value models to improve the performance further \cite{zhang2017sparse}, \cite{scheirer2014probability}. In the proposed approach as well, the tail of open-set identification scores are modeled using the extreme value distribution to find the optimal threshold for operation.


\section{Proposed Method}
\label{sec:prop_method}
The proposed approach divides the open-set recognition problem into two sub-tasks, namely, closed-set classification and open-set identification. The training procedure for these tasks are shown in Fig.~\ref{fig:block_dia} as stage-1 and stage-2. Stage-3 in Fig.~\ref{fig:block_dia} provides overview of the proposed approach at inference. In what follows, we present details of these stages.

\subsection{Closed-set Training (Stage 1)}
\label{subsec:s1_cc}


Given images in a batch $\{X_1, X_2,..., X_N \} \in \mathcal{K}$, and their corresponding labels $\{y_1, y_2,..., y_N \}$. Here $N$ is the batch size and $\forall y_i\in\{1,2,..,k\}$. The encoder ($\mathcal{F}$) and the classifier ($\mathcal{C}$) with parameters $\Theta_f$ and $\Theta_c$, respectively are trained using the following cross entropy loss,
\vskip -10.0pt
\begin{equation}
\begin{aligned}
\mathcal{L}_c(\{\Theta_f, \Theta_c\}) \ = \ -\frac{1}{N} \sum_{i=1}^{N} 
\sum_{j=1}^{k} \ \mathbb{I}_{y_i}(j) \  \text{log} [ p_{y_i}(j) ],
\end{aligned}
\label{eq:lc}
\end{equation}

where, $\mathbb{I}_{y_i}$ is an indicator function for label $y_i$ (\textit{i.e.}, one hot encoded vector) and $p_{y_i} = \mathcal{C}(\mathcal{F}(X_i))$ is a predicted probability score vector. $p_{y_i}(j)$ is probability of the $i^{th}$ sample being from the $j^{th}$ class.

\subsection{Open-set Training (Stage 2)}
\label{subsec:s2_oi}

There are two major parts in open-set training, conditional decoder training, followed by EVT modeling of the reconstruction errors. In this stage, the encoder and classifier weights are fixed and don't change during optimization.

\subsubsection{Conditional Decoder Training}
\label{subsubsec:s2_dec}

\begin{figure*}[!t]
	\centering
	\begin{subfigure}[t]{0.5\textwidth}
		\centering
		\includegraphics[width=1.0\linewidth]{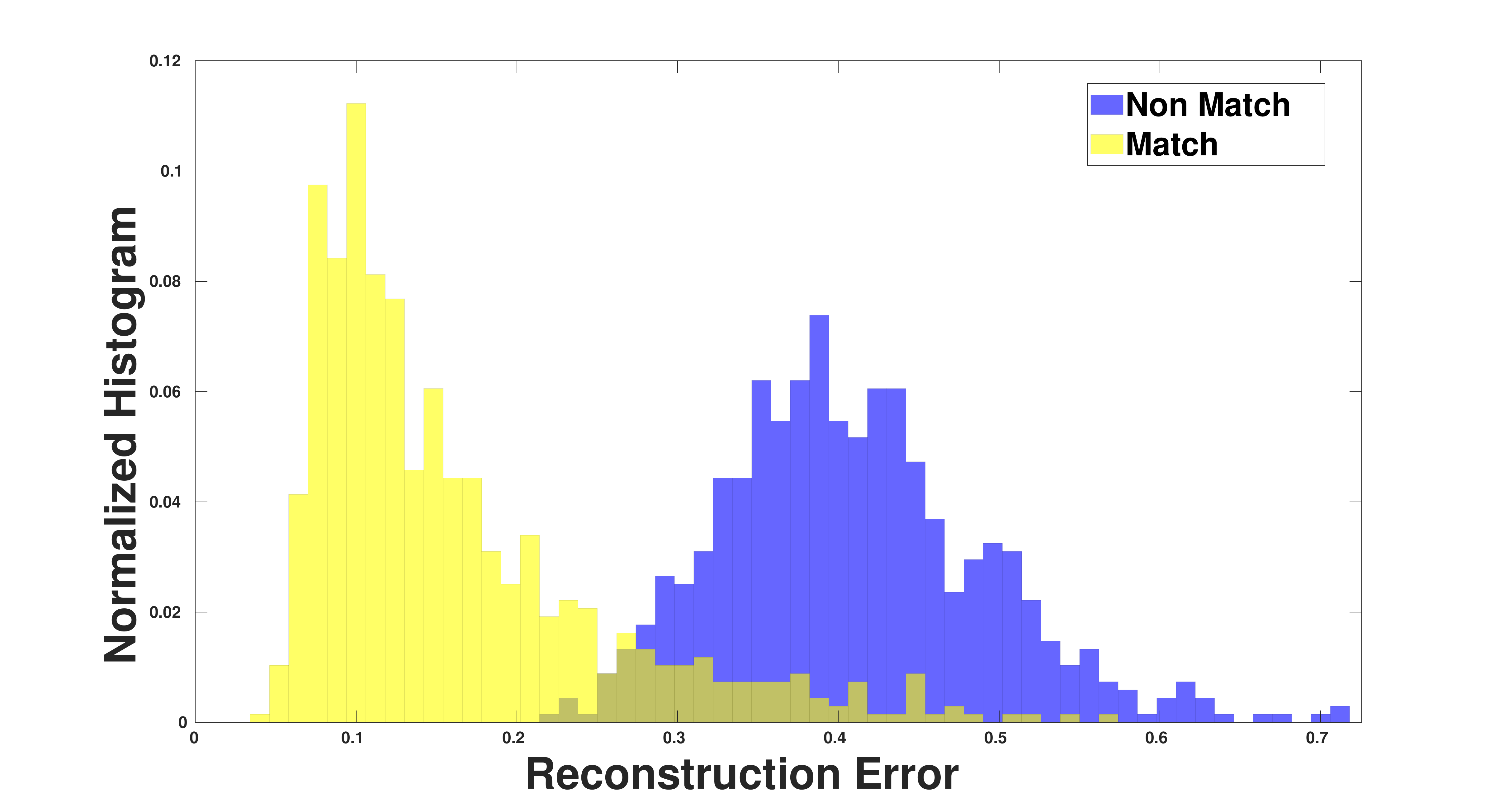}
		\vskip -7.5pt \caption{Normalized histogram of match and non-match reconstruction  errors.}
		\label{fig:svhn_a}
	\end{subfigure}%
	~
	\begin{subfigure}[t]{0.5\textwidth}
		\centering
		\includegraphics[width=1.0\linewidth]{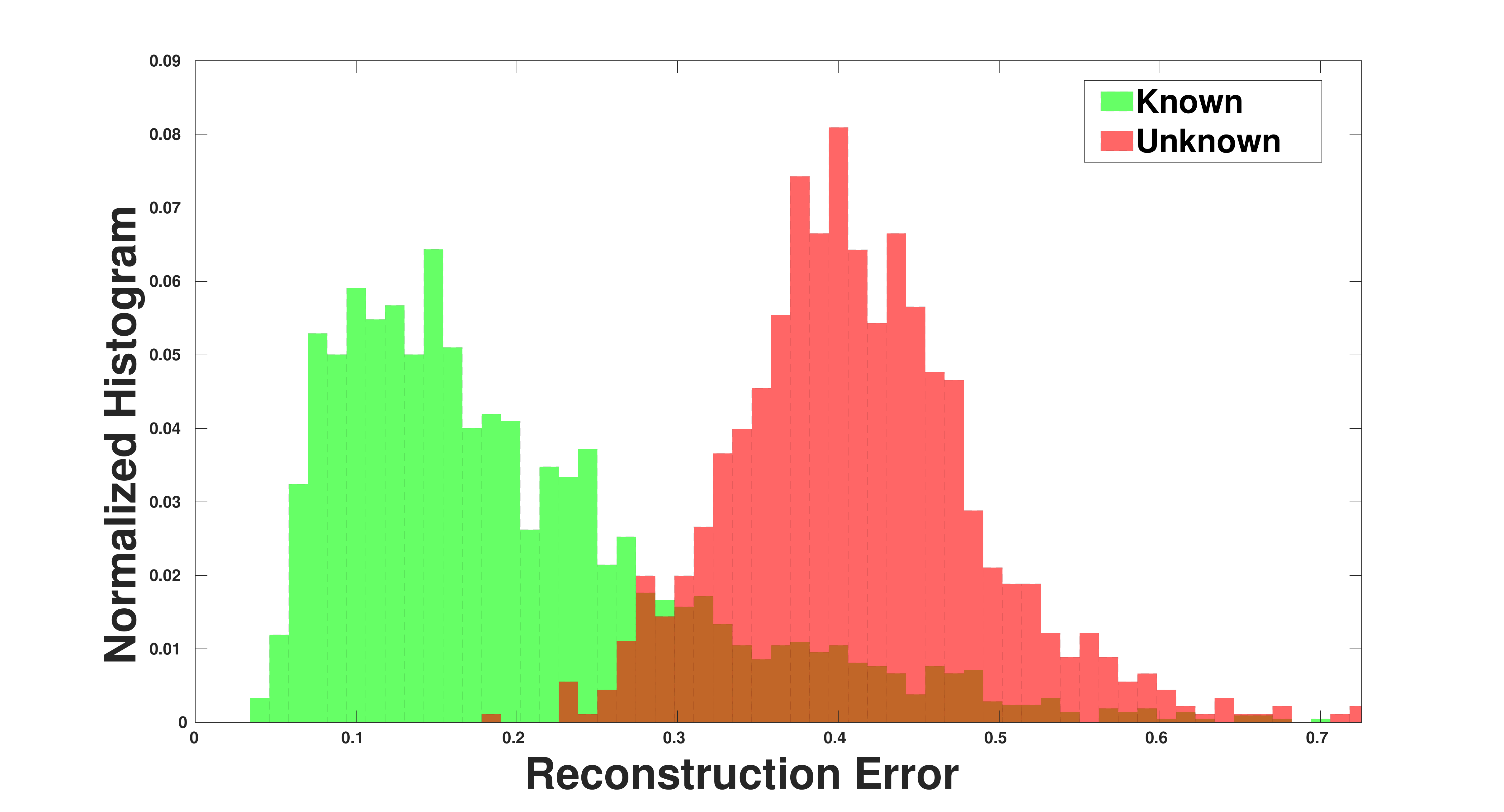}
		\vskip -7.5pt \caption{Normalized histogram of known and unknown reconstruction errors.}
		\label{fig:svhn_b}
	\end{subfigure}    
	
	\vskip -10.0pt \caption{Histogram of the reconstruction errors corresponding to the SVHN dataset.}
	\label{fig:svhn}
\end{figure*}

For any batch described in Sec.~\ref{subsec:s1_cc}, $\mathcal{F}$ is used to extract the latent vectors as, $\{z_1,z_2,...,z_N\}$. This latent vector batch is conditioned following the work by Perez \emph{et al.} \cite{perez2017film} called FiLM. FiLM influences the input feature map by applying a feature-wise linear modulations (hence the name FiLM) based on conditioning information. For a input feature $z$ and vector $l_j$ containing conditioning information can be given as,
\vskip -17.5pt
\begin{flalign}
\gamma_j &= H_{\gamma}(l_j), \ \ \ \beta_j = H_{\beta}(l_j), \\
z_{l_j}  &= \gamma_j \odot z + \beta_j,
\label{eq:film_master}
\end{flalign}
\vskip -10.0pt
where,
\vskip -20.0pt
$$ l_j(x)=
\begin{cases}
+1, x = j,\\
-1,x \neq j,\\
\end{cases}
x,j \ \in \{1, 2,..., k\}.
$$
Here, $H_{\gamma}$ and $H_{\beta}$ are neural networks with parameters $\Theta_{\gamma}$ and $\Theta_{\beta}$. Tensors $z_{l_j}$, $\gamma_{j}$, $\beta_{j}$ have the same shape and $\odot$ represents the Hadamard product. $l_j$ is used for conditioning, and referred to as label condition vector in the paper. Also, the notation $z_{l_j}$ is used to describe the latent vector $z$ conditioned on the label condition vector $l_j$, i.e,  $z|{l_j}$.

The decoder ($\mathcal{G}$ with parameters $\Theta_g$) is expected to perfectly reconstruct the original input when conditioned on the label condition vector matching the class identity of the input, referred here as the match condition vector ($l_{m}$), can be viewed as a traditional auto-encoder. However, here $\mathcal{G}$ is additionally trained to poorly reconstruct the original input when conditioned on the label condition vector, that does not match the class identity of the input, referred here as the non-match condition vector ($l_{nm}$). The importance of this additional constraint on the decoder is discussed in Sec.~\ref{subsubsec:s2_evt} while modeling the reconstruction errors using EVT. For the rest of this paper, we use superscript $m$ and $nm$ to indicate match and non-match, respectively.


Now, for a given input $X_i$ from the batch and $l_m = l_{y^{m}_i}$ and $l_{nm} = l_{y_j^{nm}}$, for any random $y^{nm}_i \neq y_i$ sampled from $\{1, 2,..,k\}$, be its corresponding match and non-match condition vectors, the feed forward path for stage-2 can be summarized through the following equations,
\vskip -17.5pt
\begin{flalign*}
z_i &= \mathcal{F}(X_i),\\
\gamma_{y^{m}_i} &= H_{\gamma}(l_{y^{m}_i}), \ \ \ \ \ \ \ \ \ \ \ \ \ \ \ \ \gamma_{y^{nm}_i} = H_{\gamma}(l_{y^{nm}_i}), \\
\beta_{y^{m}_i} &= H_{\beta}(l_{y^{m}_i}), \ \ \ \ \ \ \ \ \ \ \ \ \ \ \ \ \beta_{y^{nm}_i} = H_{\beta}(l_{y^{nm}_i}), \\
z_{il_m} &= \gamma_{y^{m}_i} \odot z_i + \beta_{y^{m}_i}, \ \ \ \ z_{il_{nm}}  = \gamma_{y^{nm}_i} \odot z_i + \beta_{y^{nm}_i}, \\
\tilde{X}^{m}_i &= \mathcal{G}(z_{l_m}). \ \ \ \ \ \ \ \ \ \ \ \ \ \ \ \ \ \ \ \tilde{X}^{nm}_i = \mathcal{G}(z_{l_{nm}}).
\label{eq:ff_s2}
\end{flalign*}

Following the above feed-forward path, the loss functions in the second stage of training to train the decoder ($\mathcal{G}$ with parameters $\Theta_g$) and conditioning layer (with parameters $\Theta_{\gamma}$ and $\Theta_{\beta}$) are given as follows,
\vskip -20.0pt
\begin{equation}
\begin{aligned}
\mathcal{L}^{m}_r(\{\Theta_g,\Theta_{\gamma},\Theta_{\beta}\}) = \frac{1}{N} \sum_{i=1}^{N} ||X_i-\tilde{X}^{m}_i||_1, 
\end{aligned}
\label{eq:lr_m}
\end{equation}

\vskip -17.5pt
\begin{equation}
\begin{aligned}
\mathcal{L}^{nm}_r(\{\Theta_g,\Theta_{\gamma},\Theta_{\beta}\}) = \frac{1}{N} \sum_{i=1}^{N} ||X^{nm}_i-\tilde{X}_i^{nm}||_1, 
\end{aligned}
\label{eq:lr_nm}
\end{equation}


\vskip -12.5pt
\begin{equation}
\begin{aligned}
\underset{\{ \Theta_g,\Theta_{\gamma},\Theta_{\beta} \}}{\text{min}}
\alpha\mathcal{L}_r^{m}(\{ \Theta_g,\Theta_{\gamma},\Theta_{\beta} \}) \\[-1.00em]+(1-\alpha)&\mathcal{L}_r^{nm}(\{ \Theta_g,\Theta_{\gamma},\Theta_{\beta} \}).
\end{aligned}
\label{eq:lr_full}
\end{equation}

Here, the loss function $\mathcal{L}^{m}_r$ corresponds to the constraint that output generated using match condition vector $\tilde{X}^{m}_i$, should be perfect reconstruction of $X_i$. Whereas, the loss function $\mathcal{L}^{nm}_r$ corresponds to the constraint that output generated using non match condition vector $\tilde{X}^{nm}_i$, should have poor reconstruction. To enforce the later condition, another batch $\{X_1^{nm}, X_2^{nm},..., X_N^{nm} \}$, is sampled from the training data, such that new batch does not have class identity consistent with the match condition vector. This in effect achieves the goal of poor reconstruction when conditioned $l_{y^{nm}_i}$. This conditioning strategy in a way, emulates open-set behavior (as will be discussed further in Sec.~\ref{subsubsec:s2_evt}). Here, the network is specifically trained to produce poor reconstructions when class identity of an input image does not match the condition vector. So, when encountered with an unknown class test sample, ideally none of the condition vector would match the input image class identity. This will result in poor reconstruction for all condition vectors. While, when encountered with the known test sample, as one of the condition vector will match the input image class identity, it will produce a perfect reconstruction for that particular condition vector. Hence, training with the non-match loss helps the network adapt better to open-set setting. Here, $\mathcal{L}^{nm}_r$ and $\mathcal{L}^{m}_r$ are weighted with $\alpha \in [0, 1]$.

\subsubsection{EVT Modeling}
\label{subsubsec:s2_evt}
\textbf{Extreme Value Theory.} Extreme value theory is often used in many visual recognition systems and is an effective tool for modeling post-training scores \cite{scheirer2014probability}, \cite{shi2008modeling}.  It has been used in many applications such as finance, railway track inspection etc. \cite{evtweather}, \cite{evtfinance}, \cite{evtrailway} as well as open-set recognition \cite{bendale2016towards}, \cite{scheirer2014probability}, \cite{zhang2017sparse}. In this paper we follow the Picklands-Balkema-deHaan formulation \cite{pickands1975statistical}, \cite{balkema1974residual} of the extreme value theorem. It considers modeling probabilities conditioned on random variable exceeding a high threshold. For a given random variable $W$ with a cumulative distribution function (CDF) $F_W(w)$ the conditional CDF for any $w$ exceeding the threshold $u$ is defined as,
\vskip -15.0pt
\begin{equation*}
F_U(w) = \mathcal{P}(w - u \leq w | w > u) = \frac{F_W(u+w)-F_W(u)}{1-F_W(u)}
,\label{eq:gev}
\end{equation*}
where, $\mathcal{P}(\cdot)$ denotes probability measure function. Now, given I.I.D. samples, $\{ W_i,..., W_n\}$, the extreme value theorem \cite{pickands1975statistical} states that, for large class of underlying distributions and given a large enough $u$, $F_U$ can be well approximated by the Generalized Pareto Distribution (GPD),
\vskip -15.0pt
\begin{equation}
G(w;\zeta, \mu)=
\begin{cases}
1 - (1+ \frac{\zeta \cdot w}{\mu})^{\frac{1}{\zeta}}, \ \text{if} \ \zeta \neq 0,\\
1 - e^{\frac{w}{\mu}} \ \ \ \ \ \ \ \ \ \ \ \ \ , \ \text{if} \ \zeta = 0,\\
\end{cases}
\end{equation}
such that $-\infty < \zeta < +\infty$, $0 < \mu < +\infty$, $w > 0$ and $\zeta w > -\mu$. $G(.)$ is CDF of GPD and for $\zeta=0$, reduces to the exponential distribution with parameter $\mu$ and for $\zeta \neq 0$ takes the form of Pareto distribution \cite{gpd}.

\textbf{Parameter Estimation.} When modeling the tail of any distribution as GPD, the main challenge is in finding the tail parameter $u$ to get the conditional CDF. However, it is possible to find an estimated value of $u$ using mean excess function (MEF), \textit{i.e.}, $E[W-u|W>u]$ \cite{shi2008modeling}. It has been shown that for GPD, MEF holds a linear relationship with $u$. Many researchers use this property of GPD to estimate the value of $u$ \cite{shi2008modeling}, \cite{perera2017extreme}. Here, the algorithm for finding $u$, introduced in \cite{perera2017extreme} for GPD is adopted with minor modifications. See \cite{perera2017extreme}, \cite{shi2008modeling} for  more details regarding MEF or tail parameter estimation.    After getting an estimate for $u$, since from extreme value theorem \cite{pickands1975statistical}, we know that set $\{w \in W \ | \ w > u\}$, follows GPD distribution, rest of the parameters for GPD, \textit{i.e.} $\zeta$ and $\mu$ can be easily estimated using the maximum likelihood estimation techniques \cite{grimshaw1993computing}, except for some rarely observed cases \cite{choulakian2001goodness}.

\subsubsection{Threshold Calculation}

After training procedure described in previous sections, Sec.~\ref{subsec:s1_cc} and Sec.~\ref{subsec:s2_oi}, set of match and non-match reconstruction errors are created from training set, $\{X_1, X_2,..., X_{N_{train}} \} \in \mathcal{K}$, and their corresponding match and non match labels, $\{y^{m}_1, y^{m}_2,..., y^{m}_{N_{train}} \}$ and $\{y_1^{nm}, y_2^{nm},..., y_{N_{train}}^{nm} \}$. Let, $r^m_i$ be the match reconstruction error and $r^{nm}_i$ be the non match reconstruction error for the input $X_i$, then the set of match and non match errors can be calculated as,

\vskip -00.0pt
\begin{alignat*}{2}
\tilde{X}^{m}_i &=\mathcal{G}(\mathcal{H}_{\gamma}(l_{y^{m}_i})\odot \mathcal{F}(X_i) + \mathcal{H}_{\beta}(l_{y^{m}_i}) ),\\
\tilde{X}^{nm}_i &=\mathcal{G}(\mathcal{H}_{\gamma}(l_{y^{nm}_i})\odot \mathcal{F}(X_i)+\mathcal{H}_{\beta}(l_{y^{nm}_i})),
\end{alignat*}

\vskip -20.0pt
\begin{alignat*}{3}
S_m &= \ \{ r^m \in \mathbb{R}^+ \cup \{0\} \ | \ \ r_i^m = ||X_i-\tilde{X}^{m}_i||_1 \  \},\\
S_{nm} &= \ \{ r^{nm} \in \mathbb{R}^+  \cup \{0\} \ | \  \ r_i^{nm} = ||X_i-\tilde{X}^{nm}_i||_1 \  \},\\
\forall i &\in \{1,2,...,N_{train}\}.
\end{alignat*}

Typical histograms of $S_m$ (set of match reconstruction errors) and $S_{nm}$ (set of non-match reconstruction errors) are shown in Fig.~\ref{fig:svhn_a}.  Note that the elements in these sets are calculated solely based on what is observed during training (i.e., without utilizing any unknown samples). Fig.~\ref{fig:svhn_b} shows the normalized histogram of the reconstruction errors observed during inference from the test samples of known class set $(\mathcal{K})$, and unknown class set $(\mathcal{U})$. Comparing these figures in Fig.~\ref{fig:svhn}, it can be observed that the distribution of $S_m$ and $S_{nm}$ computed during training, provides a good approximation for the error distributions observed during inference, for test samples from known set $(\mathcal{K})$ and unknown set $(\mathcal{U})$. This observation also validates that non match training emulates an open-set test scenario (also discussed in Sec.~\ref{subsec:s2_oi}) where the input does not match any of the class labels. This motivates us to use $S_m$ and $S_{nm}$ to find an operating threshold for open-set recognition to make a decision about any test sample being known/unknown.

Now, It is safe to assume that the optimal operating threshold ($\tau^*$) lies in the region $S_m \cap S_{nm}$. Here, the underlying distributions of $S_m$ and $S_{nm}$ are not known. However, as explained in \ref{subsubsec:s2_evt}, it is possible to model the tails of $S_m$ (right tail) and $S_{nm}$ (left tail) with GPD as $G_m$ and $G_{nm}$ with $G(\cdot)$ being a CDF. Though, GPD is only defined for modeling maxima, before fitting $G_{nm}$ left tail of $S_{nm}$ we perform inverse transform as $S_{nm}'=-S_{nm}$. Assuming the prior probability of observing unknown samples is $p_u$, the probability of errors can be formulated as a function of the threshold $\tau$,

\vskip -00.0pt
\begin{equation*}
\begin{aligned}
& \tau^* = \underset{\tau}{\text{min}}
& & \mathcal{P}_{error}(\tau)\\
& \ \ \ \ = \underset{\tau}{\text{min}}
& & [ (1-p_u) * \mathcal{P}_{m}(r>\tau) + p_u * \mathcal{P}_{nm}(-r<-\tau) ]\\
& \ \ \ \ = \underset{\tau}{\text{min}}
& & [ (1-p_u) *(1-G_{m}(\tau)) + p_u * (1-G_{nm}(\tau)) ].\\
\end{aligned}
\end{equation*}
\vskip -0.0pt

Solving the above equation should give us an operating threshold that can minimize the probability of errors for a given model and can be solved by a simple line search algorithm by searching for $\tau^*$ in the range $\{S_m \cap S_{nm}\}$. Here, the accurate estimation of $\tau^*$ depends on how well $S_m$ and $S_{nm}$  represent the known and unknown error distributions. It also depends on the prior probability $p_u$, effect of this prior will be further discussed in Sec.~\ref{subsec:exp_2}.

\subsection{Open-set Testing by k-inference (Stage 3)}
\label{subsec:s3_kinf}

Here, we introduce the open-set testing algorithm for proposed method. The testing procedure is described in Algo.~\ref{algo:k_inf} and an overview of this is also shown in Fig.~\ref{fig:block_dia}. This testing strategy involves conditioning the decoder $k$-times with all possible condition vectors to get $k$ reconstruction errors. Hence, it is referred as $k$-inference algorithm.


\section{Experiments and Results}
\label{sec:exp_res}

In this section we evaluate the performance of the proposed approach and compare it  with the state of the art open-set recognition methods. The experiments in Sec.~\ref{subsec:exp_1}, we measure the ability of algorithm to identify test samples as known or unknown without considering operating threshold. In second set of experiments in Sec.~\ref{subsec:exp_2}, we measure overall performance (evaluated using F-measure) of open-set recognition algorithm. Additionally through ablation experiments, we analyze contribution from each component of the proposed method.

\subsection{Implementation Details}
\label{subsec:impl}

\begin{algorithm}[t]
	\caption{k-Inference Algorithm}\label{algo:k_inf}
	\begin{algorithmic}[1]
		\Require Trained network models $\mathcal{F}$, $\mathcal{C}$, $\mathcal{G}$, $\mathcal{H}_{\gamma}$, $\mathcal{H}_{\beta}$
		\Require Threshold $\tau$ from EVT model
		\Require Test image $X$, $k$ condition vectors $\{l_1, \ . \ . \ . \ , l_k\}$ 
		\State Latent space representation, $z = \mathcal{F}(X)$
		\State Prediction probabilities,  $p_{y} = \mathcal{C}(z)$
		\State predict known label,  $y_{pred} = argmax(p_y)$
		\State \textbf{for} $i=1, \ . \ . \ . \ ,k$ \textbf{do}
		\State $ \ \ \ \ \ \ \ \ \ \ z_{l_{i}} = \mathcal{H}_{\gamma}(l_{i})\odot z + \mathcal{H}_{\beta}(l_{i})$
		\State $ \ \ \ \ \ \ \ \ \ \ \tilde{X}_i = \mathcal{G}(z_{l_{i}})$
		\State $ \ \ \ \ \text{Rec}(i)= ||X-\tilde{X}_i||_1$
		\State \textbf{end for}
		\State $\text{Rec}_{min} = sort(\text{Rec})$
		\State \textbf{if} $ \ \ \text{Rec}_{min} \ < \ \tau \ $ \textbf{do}
		\State $ \ \ \ \ $ predict $X$ as Known, with label $y_{pred}$ 
		\State \textbf{else do}
		\State $ \ \ \ \ $ predict $X$ as Unknown
		\State \textbf{end if}
	\end{algorithmic}
\end{algorithm}

We use Adam optimizer \cite{kingma2015adam} with learning rate $0.0003$ and batch size, $N$=$64$. The parameter $\alpha$, described in Sec.~\ref{subsec:s2_oi}, is set equal to 0.9. For all the experiments, conditioning layer networks $\mathcal{H}_{\gamma}$ and $\mathcal{H}_{\beta}$ are a single layer fully connected neural networks. Another important factor affecting open-set performance is openness of the problem. Defined by Scheirer \textit{et al.} \cite{scheirer2013toward}, it quantifies how open the problem setting is,
\vskip -0.0pt
\begin{equation}
\begin{aligned}
\mathbf{O} = 1 \ - \ \sqrt{\frac{2\times N_{train}}{N_{test}+N_{target}}},
\end{aligned}
\label{eq:fm}
\end{equation}
\vskip -0.0pt
where, $N_{train}$ is the number of training classes seen during training, $N_{test}$ is the number of test classes that will be observed during testing, $N_{target}$ is the number of target classes that needs to be correctly recognized during testing. We evaluate performance over multiple openness value depending on the experiment and dataset.

\subsection{Experiment I : Open-set Identification}
\label{subsec:exp_1}

The evaluation protocol defined in \cite{neal2018open} is considered and area under ROC (AUROC) is used as evaluation metric. AUROC provides a calibration free measure and characterizes the performance for a given score by varying threshold. The encoder, decoder and classifier architecture for this experiment is similar to the architecture used by \cite{neal2018open} in their experiments. Following the protocol in \cite{neal2018open}, we report the AUROC averaged over five randomized trials.

\subsubsection{Datasets} 
\label{subsubsec:exp1_dat}

\begin{table*}[htp!]
	\centering
	\begin{tabular}{|c|c|c|c|c|c|c|}
		\hline
		\textbf{Method}   & \textbf{MNIST} & \textbf{SVHN} & \textbf{CIFAR10} & \textbf{CIFAR+10} & \textbf{CIFAR+50} & \textbf{TinyImageNet} \\ \hline
		SoftMax            & 0.978 & 0.886 & 0.677 & 0.816 & 0.805 & 0.577 \\ \hline
		OpenMax \cite{bendale2016towards} (CVPR'16)   & 0.981 & 0.894 & 0.695 & 0.817 & 0.796 & 0.576 \\ \hline
		G-OpenMax \cite{ge2017generative} (BMVC'17) & 0.984 & 0.896 & 0.675 & 0.827 & 0.819 & 0.580 \\ \hline
		OSRCI \cite{neal2018open} (ECCV'18)     & 0.988 & 0.910 & 0.699 & 0.838 & 0.827 & 0.586 \\ \hline
		Proposed Method              & \textbf{0.989} & \textbf{0.922} & \textbf{0.895} & \textbf{0.955} & \textbf{0.937} & \textbf{0.748} \\ \hline
	\end{tabular}
	\vskip -10.0pt \caption{AUROC for open-set identification, values other than the proposed method are taken from \cite{neal2018open}.}
	\label{table:auroc}
\end{table*}


Here, we provide summary of these protocols for each dataset,

\noindent \textbf{MNIST, SVHN, CIFAR10.} For MNIST \cite{lecun2010mnist}, SVHN \cite{netzer2011reading} and CIFAR10 \cite{krizhevsky2009learning}, openness of the problem is set to $\mathbf{O} = 13.39\%$, by randomly sampling 6 known classes and 4 unknown classes.\\
\textbf{CIFAR+10, CIFAR+50.} For CIFAR+$M$ experiments, 4 classes are sampled from CIFAR10 for training. $M$ non overlapping classes are used as the unknowns, which are sampled from the CIFAR100  dataset \cite{krizhevsky2009learning}. Openness of the problem for CIFAR+10 and CIFAR+50 is $\mathbf{O} = 33.33\%$ and $62.86\%$, respectively.  \\
\textbf{TinyImageNet.} For experiments with the TinyImageNet \cite{le2015tiny}, 20 known classes and 180 unknown classes with openness $\mathbf{O} = 57.35\%$ are randomly sampled for evaluation.

\subsubsection{Comparison with state-of-the-art}
\label{subsubsec:exp1_cmp_meth}

For comparing the open-set identification performance, we consider the following methods:

\noindent \textbf{I. SoftMax :} SoftMax score of a predicted class is used for open-set identification.\\
\textbf{II. OpenMax \cite{bendale2016towards}:} The score of $k$+$1^{th}$ class and score of the predicted class is used for open-set identification.\\
\textbf{III. G-OpenMax \cite{ge2017generative}:} It is a data augmentation technique, which utilizes the OpenMax scores after training the network with the generated data.\\
\textbf{IV. OSRCI \cite{neal2018open}:} Another data augmentation technique called counterfactual image generation is used for training the network for $k$+$1$ class classification. We refer to this method as Open-set Recognition using Counterfactual Images (OSRCI). The score value $P(y_{k+1})-\underset{i\leq k}{max} \ P(y_i)$ is used for open-set identification.\\

Results corresponding to this experiment are shown in Table~\ref{table:auroc}. As seen from this table, the proposed method outperform the other methods, showing that open-set identification training in proposed approach learns better scores for identifying unknown classes. From the results, we see that our method on the digits dataset produces a minor improvement compared to the other recent methods.  This is mainly do the reason that results on the digits dataset are almost saturated. On the other hand, our method performs significantly better than the other recent methods on the object datasets such as CIFAR and TinyImageNet.

\subsection{Experiment II : Open-set Recognition}
\label{subsec:exp_2}

\begin{figure*}[!t]
	\centering
	\begin{subfigure}[t]{0.5\textwidth}
		\centering
		\includegraphics[width=1.0\linewidth]{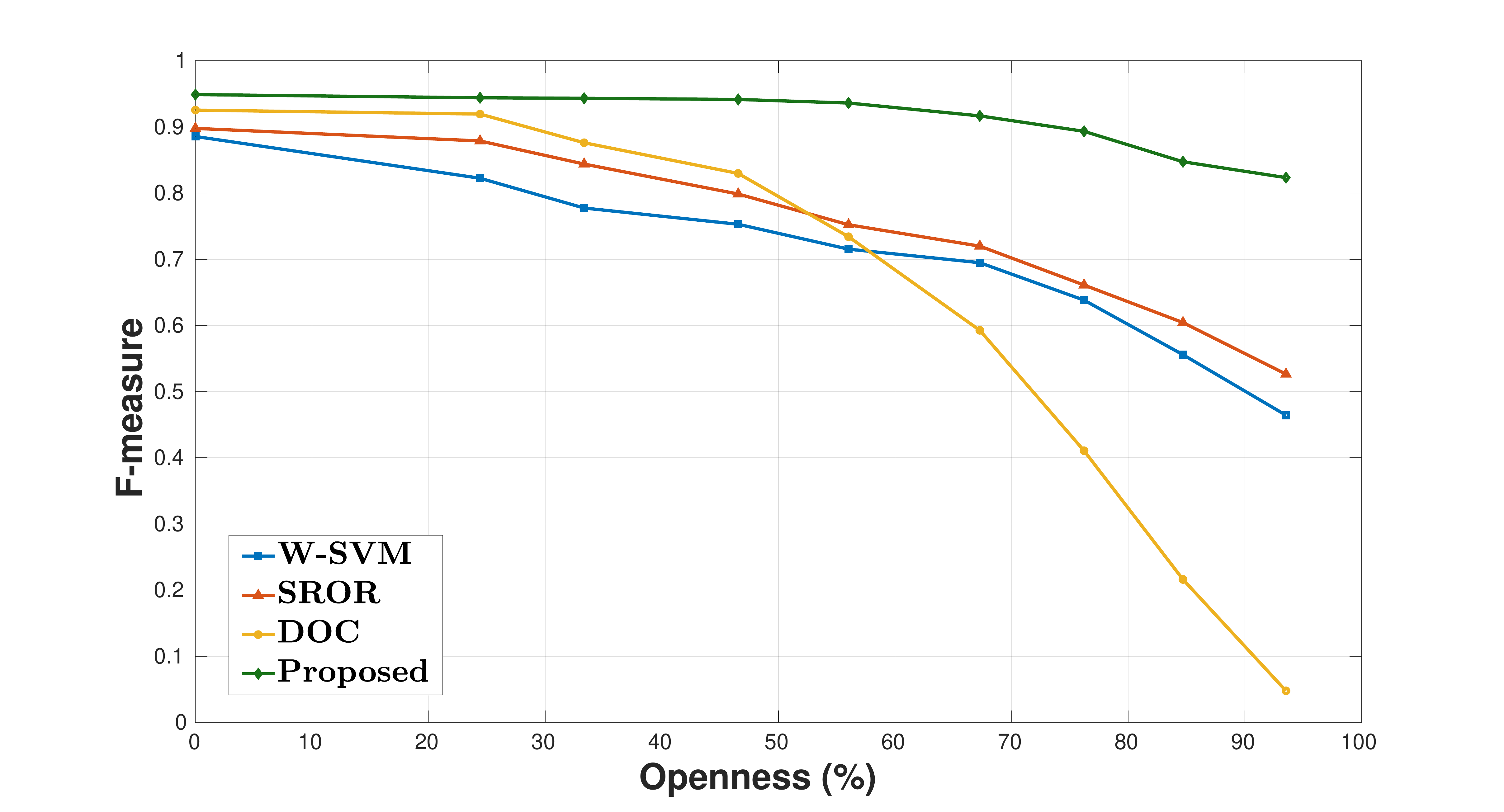}
		\vskip -8.0pt \caption{F-measure comparisons for the open-set recognition experiment.}
		\label{fig:compare_fm}
	\end{subfigure}%
	~
	\begin{subfigure}[t]{0.5\textwidth}
		\centering
		\includegraphics[width=1.0\linewidth]{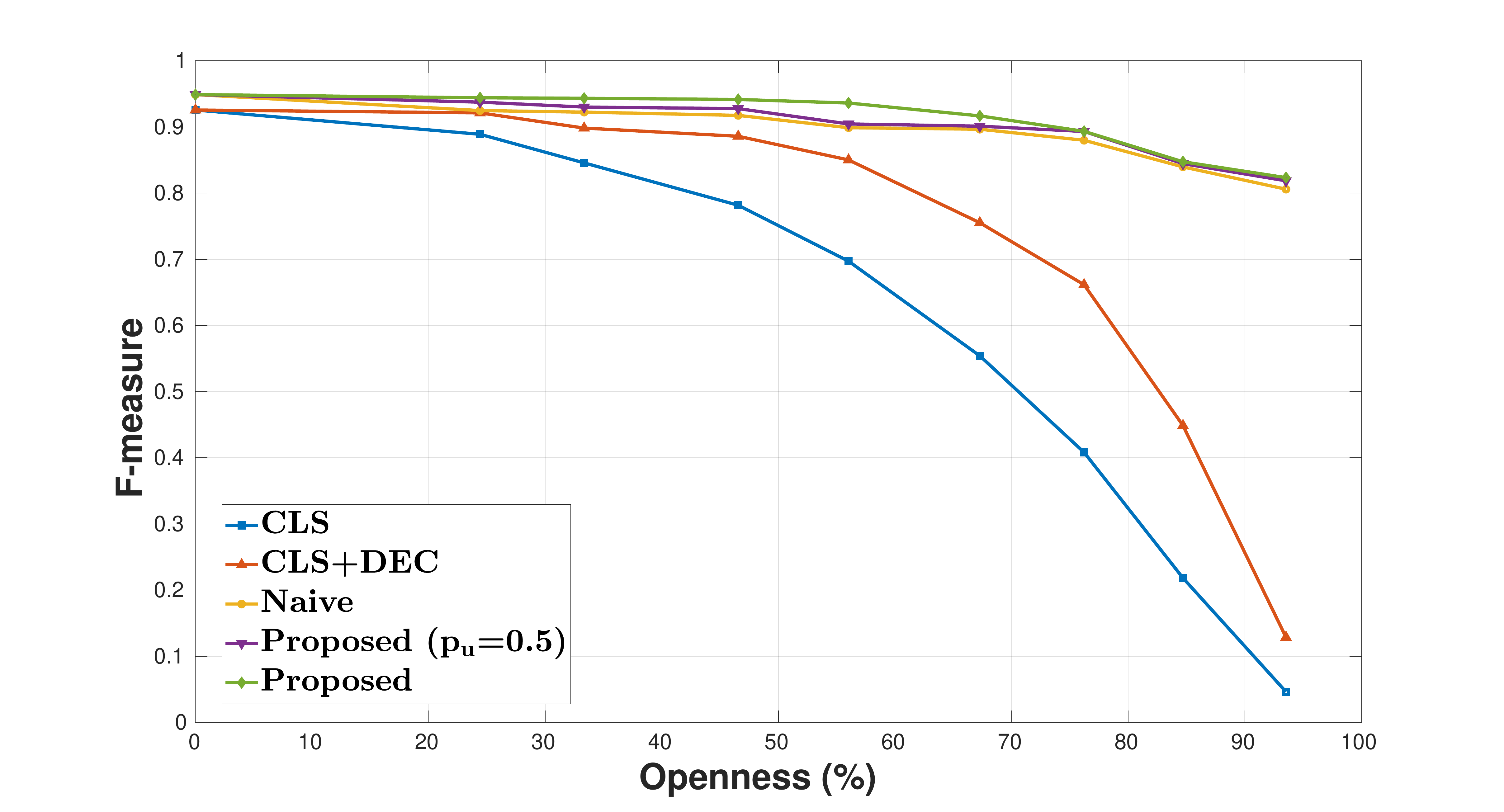}
		\vskip -10.0pt \caption{F-measure comparisons for the ablation study.}
		\label{fig:ablation_fm}
	\end{subfigure}    
	
	\vskip -10.0pt \caption{Performance evaluation on the LFW dataset.}
	\label{fig:fm_master}
\end{figure*}

This experiment shows the overall open-set recognition performance evaluated with F-measure. For this experiment we consider LFW Face dataset \cite{liu2015deep}. We extend the protocol introduced in \cite{scheirer2013toward} for open-set face recognition on LFW. Total 12 classes containing more than 50 images are considered as known classes and divided into training and testing split by 80/20 ratio. Image size is kept to 64$\times$64. Since, LFW has 5717 number of classes, we vary the openness from $0\%$ to $93\%$ by taking 18 to 5705 unknown classes during testing. Since, many classes contain only one image, instead of random sampling, we sort them according to the number of images per class and add it sequentially to increase the openness. It is obvious that with the increase in openness, the probability of observing unknown will also increase. Hence, it is reasonable to assume that prior probability $p_u$ will be a function of openness. For this experiment we set $p_u = 0.5 * \mathbf{O}$.

\subsubsection{Comparison with state-of-the-art}
\label{subsubsec:exp2_cmp_base}

For comparing the open-set recognition performance, we consider the following methods:

\noindent \textbf{I. W-SVM (PAMI'14) :} W-SVM is used as formulated in \cite{scheirer2013toward}, which trains Weibull calibrated SVM classifier for open set recognition.\\
\textbf{II. SROR (PAMI'16) :} SROR is used as formulated in \cite{zhang2017sparse}. It uses sparse representation-based framework for open-set recognition.\\
\textbf{III. DOC (EMNLP'16) :} It utilizes a novel sigmoid-based loss function for training a deep neural network \cite{shu2017doc}.

To have a fair comparison with these methods, we use features extracted from the encoder ($\mathcal{F}$) to train W-SVM and SROR.  For DOC, the encoder ($\mathcal{F}$) is trained with the loss function proposed in \cite{shu2017doc}. Experiments on LFW are performed using a U-Net \cite{ronneberger2015u} inspired encoder-decoder architecture. More details regarding network architecture is included in the supplementary material.

Results corresponding to this experiment is shown in Fig.~ \ref{fig:compare_fm}. From this figure, we can see that the proposed approach remains relatively stable with the increase in openness, outperforming all other methods. One interesting trend noticed here is, that DOC initially performs better than the statistical methods such as W-SVM and SROR. However with openness more than 50\% the performance suffers significantly. While the statistical methods though initially perform poor compared to DOC, but remain relatively stable and performs better than DOC as the openness is increased (especially over $\mathbf{O}>$50\%). 

\subsubsection{Ablation Study}
\label{subsubsec:exp2_ablation}

In this section, we present ablation analysis of the proposed approach on the LFW Face dataset. The contribution of individual components to the overall performance of the method is reported by creating multiple baselines of the proposed approach. Starting with the most simple baseline, \textit{i.e.}, thresholding SoftMax probabilities of a closed-set model, each component is added building up to the proposed approach. Detailed descriptions of these baselines are given as follows, \\
\textbf{I. CLS :} Encoder $(\mathcal{F})$ and the classifier $(\mathcal{C})$ are trained for $k$-class classification. Samples with probability score prediction less than 0.5 are classified as unknown.\\
\textbf{II. CLS+DEC :} In this baseline, only the networks $\mathcal{F}$, $\mathcal{C}$ and the decoder $(\mathcal{G})$ are trained as described in Sec.~\ref{sec:prop_method}, except $\mathcal{G}$ is only trained with match loss function, $\mathcal{L}_r^m$. Samples with more than 95\% of maximum train reconstruction error observed, are classified as unknown.\\
\textbf{III. Naive :} Here, the networks $\mathcal{F}$, $\mathcal{C}$ and $\mathcal{G}$ and the conditioning layer networks ($\mathcal{H}_{\gamma}$ and $\mathcal{H}_{\beta}$) are trained as described in Sec.~\ref{sec:prop_method}, but instead of modeling the scores using EVT as described in Sec.~\ref{subsubsec:s2_evt}, threshold is directly estimated from the raw reconstruction errors.\\  
\textbf{IV. Proposed method (\textit{\textbf{p}}$_u$ = 0.5) :} $\mathcal{F}$, $\mathcal{C}$, $\mathcal{G}$ and condition layer networks ($\mathcal{H}_{\gamma}$ and $\mathcal{H}_{\beta}$) are trained as described in Sec.~\ref{sec:prop_method} and to find the threshold prior probability of observing unknown is set to $p_u=0.5$.\\
\textbf{V. Proposed method:} Method proposed in this paper, with $p_u$ set as described in Sec.~\ref{subsec:exp_2}.

Results corresponding to the ablation study are shown in Fig.~\ref{fig:ablation_fm}. Being a simple SoftMax thresholding baseline, \textbf{CLS} has weakest performance. However, when added with a match loss function ($\mathcal{L}_r^m$) as in \textbf{CLS+DEC}, the open-set identification is performed using reconstruction scores. Since, it follows a heuristic way of thresholding, the performance degrades rapidly as openness increases. However, addition of non match loss function ($\mathcal{L}_r^{nm}$), as in the \textbf{Naive} baseline, helps find a threshold value without relying on heuristics. As seen from the Fig.~\ref{fig:ablation_fm} performance of \textbf{Naive} baseline remains relatively stable with increase in openness, showing the importance of loss function $\mathcal{L}_r^{nm}$. Proposed method with $p_u$ fixed to 0.5, introduces EVT modeling on reconstruction errors to calculate a better operating threshold. It can be seen from the Fig.~\ref{fig:ablation_fm}, such strategy improves over finding threshold based on raw score values. This shows importance applying EVT models on reconstruction errors. Now, if $p_u$ is set to $0.5*\mathbf{O}$, as in the proposed method, there is a marginal improvement over the fixed $p_u$ baseline. This shows benefit of setting $p_u$ as a function of openness. It is interesting to note that for large openness values (as $0.5*\mathbf{O} \rightarrow 0.5$), both fixed $p_u$ baseline and proposed method achieve similar performance.

\section{Conclusion}
\label{sec:con_fu}
We presented an open-set recognition algorithm based on class conditioned auto-encoders. We introduced training and testing strategy for these networks. It was also shown that dividing the open-set recognition into sub tasks helps learn a better score for open-set identification. During training, enforcing conditional reconstruction constraints are enforced, which helps learning approximate known and unknown score distributions of reconstruction errors. Later, this was used to calculate an operating threshold for the model. Since inference for a single sample needs $k$ feed-forwards, it suffers from increased test time. However, the proposed approach performs well across multiple image classification datasets and providing significant improvements over many state of the art open-set algorithms. In our future research, generative models such as GAN/VAE/FLOW can be explored to modify this method. We will revise the manuscript with such details in the conclusion.

\section*{Acknowledgements}
This research is based upon work supported by the Office of the Director of National Intelligence (ODNI), Intelligence Advanced Research Projects Activity (IARPA), via IARPA R\&D Contract No. 2014-14071600012. The views and conclusions contained herein are those of the authors and should not be interpreted as necessarily representing the official policies or endorsements, either expressed or implied, of the ODNI, IARPA, or the U.S. Government.

{\small
\bibliographystyle{ieeefullname}
\bibliography{egbib}

\begin{thebibliography}{10}\itemsep=-1pt

\bibitem{evtfinance}
Isabel~Fraga Alves and Cl{\'a}udia Neves.
\newblock Extreme value distributions.
\newblock In {\em International encyclopedia of statistical science}, pages
  493--496. Springer, 2011.

\bibitem{balkema1974residual}
August~A Balkema and Laurens De~Haan.
\newblock Residual life time at great age.
\newblock {\em The Annals of probability}, pages 792--804, 1974.

\bibitem{bendale2015towards}
Abhijit Bendale and Terrance Boult.
\newblock Towards open world recognition.
\newblock In {\em Proceedings of the IEEE Conference on Computer Vision and
  Pattern Recognition}, pages 1893--1902, 2015.

\bibitem{bendale2016towards}
Abhijit Bendale and Terrance~E Boult.
\newblock Towards open set deep networks.
\newblock In {\em Proceedings of the IEEE conference on computer vision and
  pattern recognition}, pages 1563--1572, 2016.

\bibitem{choulakian2001goodness}
Vartan Choulakian and Michael~A Stephens.
\newblock Goodness-of-fit tests for the generalized pareto distribution.
\newblock {\em Technometrics}, 43(4):478--484, 2001.

\bibitem{gpd}
Herbert~Aron David and Haikady~Navada Nagaraja.
\newblock {\em Order statistics}.
\newblock Wiley Online Library, 1970.

\bibitem{dhamija2018reducing}
Akshay~Raj Dhamija, Manuel G{\"u}nther, and Terrance Boult.
\newblock Reducing network agnostophobia.
\newblock In {\em Advances in Neural Information Processing Systems}, pages
  9175--9186, 2018.

\bibitem{ge2017generative}
ZongYuan Ge, Sergey Demyanov, Zetao Chen, and Rahil Garnavi.
\newblock Generative openmax for multi-class open set classification.
\newblock {\em arXiv preprint arXiv:1707.07418}, 2017.

\bibitem{evtrailway}
Xavier Gibert, Vishal~M Patel, and Rama Chellappa.
\newblock Deep multitask learning for railway track inspection.
\newblock {\em IEEE Transactions on Intelligent Transportation Systems},
  18(1):153--164, 2017.

\bibitem{goodfellow2014generative}
Ian Goodfellow, Jean Pouget-Abadie, Mehdi Mirza, Bing Xu, David Warde-Farley,
  Sherjil Ozair, Aaron Courville, and Yoshua Bengio.
\newblock Generative adversarial nets.
\newblock In {\em Advances in neural information processing systems}, pages
  2672--2680, 2014.

\bibitem{grimshaw1993computing}
Scott~D Grimshaw.
\newblock Computing maximum likelihood estimates for the generalized pareto
  distribution.
\newblock {\em Technometrics}, 35(2):185--191, 1993.

\bibitem{he2015delving}
Kaiming He, Xiangyu Zhang, Shaoqing Ren, and Jian Sun.
\newblock Delving deep into rectifiers: Surpassing human-level performance on
  imagenet classification.
\newblock In {\em Proceedings of the IEEE international conference on computer
  vision}, pages 1026--1034, 2015.

\bibitem{resnet16}
Kaiming He, Xiangyu Zhang, Shaoqing Ren, and Jian Sun.
\newblock Deep residual learning for image recognition.
\newblock In {\em Proceedings of the IEEE conference on computer vision and
  pattern recognition}, pages 770--778, 2016.

\bibitem{senet17}
Jie Hu, Li Shen, and Gang Sun.
\newblock Squeeze-and-excitation networks.
\newblock {\em arXiv preprint arXiv:1709.01507}, 2017.

\bibitem{junior2016specialized}
Pedro Ribeiro~Mendes J{\'u}nior, Terrance~E Boult, Jacques Wainer, and Anderson
  Rocha.
\newblock Specialized support vector machines for open-set recognition.
\newblock {\em arXiv preprint arXiv:1606.03802}, 2016.

\bibitem{junior2017nearest}
Pedro R~Mendes J{\'u}nior, Roberto~M de Souza, Rafael de~O Werneck, Bernardo~V
  Stein, Daniel~V Pazinato, Waldir~R de Almeida, Ot{\'a}vio~AB Penatti, Ricardo
  da~S Torres, and Anderson Rocha.
\newblock Nearest neighbors distance ratio open-set classifier.
\newblock {\em Machine Learning}, 106(3):359--386, 2017.

\bibitem{kingma2015adam}
Diederik~P Kingma and Jimmy Ba.
\newblock Adam: A method for stochastic optimization.
\newblock 2015.

\bibitem{krizhevsky2009learning}
Alex Krizhevsky.
\newblock Learning multiple layers of features from tiny images.
\newblock Technical report, Citeseer, 2009.

\bibitem{alexnet12}
Alex Krizhevsky, Ilya Sutskever, and Geoffrey~E Hinton.
\newblock Imagenet classification with deep convolutional neural networks.
\newblock In {\em Advances in neural information processing systems}, pages
  1097--1105, 2012.

\bibitem{le2015tiny}
Ya Le and Xuan Yang.
\newblock Tiny imagenet visual recognition challenge.
\newblock {\em CS 231N}, 2015.

\bibitem{lecun2010mnist}
Yann LeCun, Corinna Cortes, and CJ Burges.
\newblock Mnist handwritten digit database.
\newblock {\em AT\&T Labs [Online]. Available: http://yann. lecun.
  com/exdb/mnist}, 2, 2010.

\bibitem{liu2015deep}
Ziwei Liu, Ping Luo, Xiaogang Wang, and Xiaoou Tang.
\newblock Deep learning face attributes in the wild.
\newblock In {\em Proceedings of the IEEE International Conference on Computer
  Vision}, pages 3730--3738, 2015.

\bibitem{evtweather}
Gerald~A Meehl, Thomas Karl, David~R Easterling, Stanley Changnon, Roger
  Pielke~Jr, David Changnon, Jenni Evans, Pavel~Ya Groisman, Thomas~R Knutson,
  Kenneth~E Kunkel, et~al.
\newblock An introduction to trends in extreme weather and climate events:
  observations, socioeconomic impacts, terrestrial ecological impacts, and
  model projections.
\newblock {\em Bulletin of the American Meteorological Society},
  81(3):413--416, 2000.

\bibitem{neal2018open}
Lawrence Neal, Matthew Olson, Xiaoli Fern, Weng-Keen Wong, and Fuxin Li.
\newblock Open set learning with counterfactual images.
\newblock In {\em Proceedings of the European Conference on Computer Vision
  (ECCV)}, pages 613--628, 2018.

\bibitem{netzer2011reading}
Yuval Netzer, Tao Wang, Adam Coates, Alessandro Bissacco, Bo Wu, and Andrew~Y
  Ng.
\newblock Reading digits in natural images with unsupervised feature learning.
\newblock In {\em NIPS workshop on deep learning and unsupervised feature
  learning}, volume 2011, page~5, 2011.

\bibitem{oza2019active}
Poojan Oza and Vishal~M Patel.
\newblock Active authentication using an autoencoder regularized cnn-based
  one-class classifier.
\newblock {\em arXiv preprint arXiv:1903.01031}, 2019.

\bibitem{oza2019one}
Poojan Oza and Vishal~M Patel.
\newblock One-class convolutional neural network.
\newblock {\em IEEE Signal Processing Letters}, 26(2):277--281, 2019.

\bibitem{perera2019ocgan}
Pramuditha Perera, Ramesh Nallapati, and Bing Xiang.
\newblock Ocgan: One-class novelty detection using gans with constrained latent
  representations.
\newblock {\em arXiv preprint arXiv:1903.08550}, 2019.

\bibitem{perera2017extreme}
Pramuditha Perera and Vishal~M Patel.
\newblock Extreme value analysis for mobile active user authentication.
\newblock In {\em 2017 12th IEEE International Conference on Automatic Face \&
  Gesture Recognition (FG 2017)}, pages 346--353. IEEE, 2017.

\bibitem{perera2018learning}
Pramuditha Perera and Vishal~M Patel.
\newblock Learning deep features for one-class classification.
\newblock {\em arXiv preprint arXiv:1801.05365}, 2018.

\bibitem{perera2019deep}
Pramuditha Perera and Vishal~M Patel.
\newblock Deep transfer learning for multiple class novelty detection.
\newblock {\em arXiv preprint arXiv:1903.02196}, 2019.

\bibitem{perez2017film}
Ethan Perez, Florian Strub, Harm De~Vries, Vincent Dumoulin, and Aaron
  Courville.
\newblock Film: Visual reasoning with a general conditioning layer.
\newblock {\em arXiv preprint arXiv:1709.07871}, 2017.

\bibitem{pickands1975statistical}
James Pickands~III et~al.
\newblock Statistical inference using extreme order statistics.
\newblock {\em the Annals of Statistics}, 3(1):119--131, 1975.

\bibitem{ronneberger2015u}
Olaf Ronneberger, Philipp Fischer, and Thomas Brox.
\newblock U-net: Convolutional networks for biomedical image segmentation.
\newblock In {\em International Conference on Medical image computing and
  computer-assisted intervention}, pages 234--241. Springer, 2015.

\bibitem{scheirer2013toward}
Walter~J Scheirer, Anderson de Rezende~Rocha, Archana Sapkota, and Terrance~E
  Boult.
\newblock Toward open set recognition.
\newblock {\em IEEE transactions on pattern analysis and machine intelligence},
  35(7):1757--1772, 2013.

\bibitem{scheirer2014probability}
Walter~J Scheirer, Lalit~P Jain, and Terrance~E Boult.
\newblock Probability models for open set recognition.
\newblock {\em IEEE transactions on pattern analysis and machine intelligence},
  36(11):2317--2324, 2014.

\bibitem{shi2008modeling}
Zhixin Shi, Frederick Kiefer, John Schneider, and Venu Govindaraju.
\newblock Modeling biometric systems using the general pareto distribution
  (gpd).
\newblock In {\em Biometric Technology for Human Identification V}, volume
  6944, page 69440O. International Society for Optics and Photonics, 2008.

\bibitem{shu2017doc}
Lei Shu, Hu Xu, and Bing Liu.
\newblock Doc: Deep open classification of text documents.
\newblock {\em arXiv preprint arXiv:1709.08716}, 2017.

\bibitem{simonyan2014very}
Karen Simonyan and Andrew Zisserman.
\newblock Very deep convolutional networks for large-scale image recognition.
\newblock {\em arXiv preprint arXiv:1409.1556}, 2014.

\bibitem{zhang2017sparse}
He Zhang and Vishal~M Patel.
\newblock Sparse representation-based open set recognition.
\newblock {\em IEEE transactions on pattern analysis and machine intelligence},
  39(8):1690--1696, 2017.

\end{thebibliography}
}

\onecolumn

\section{Supplementary Material for C2AE: Class Conditioned Auto-Encoder for Open-set Recognition}

This contains the supplementary material for the paper  \textit{C2AE: Class Conditioned Auto-Encoder for Open-set Recognition}. Due to the space limitations  in the submitted paper, we provide some additional details regarding the proposed method.

\subsection{Toy Examples}

\begin{figure*}[htp!]
	\centering
	\begin{subfigure}[t]{0.33\textwidth}
		\centering
		\includegraphics[width=1.0\linewidth]{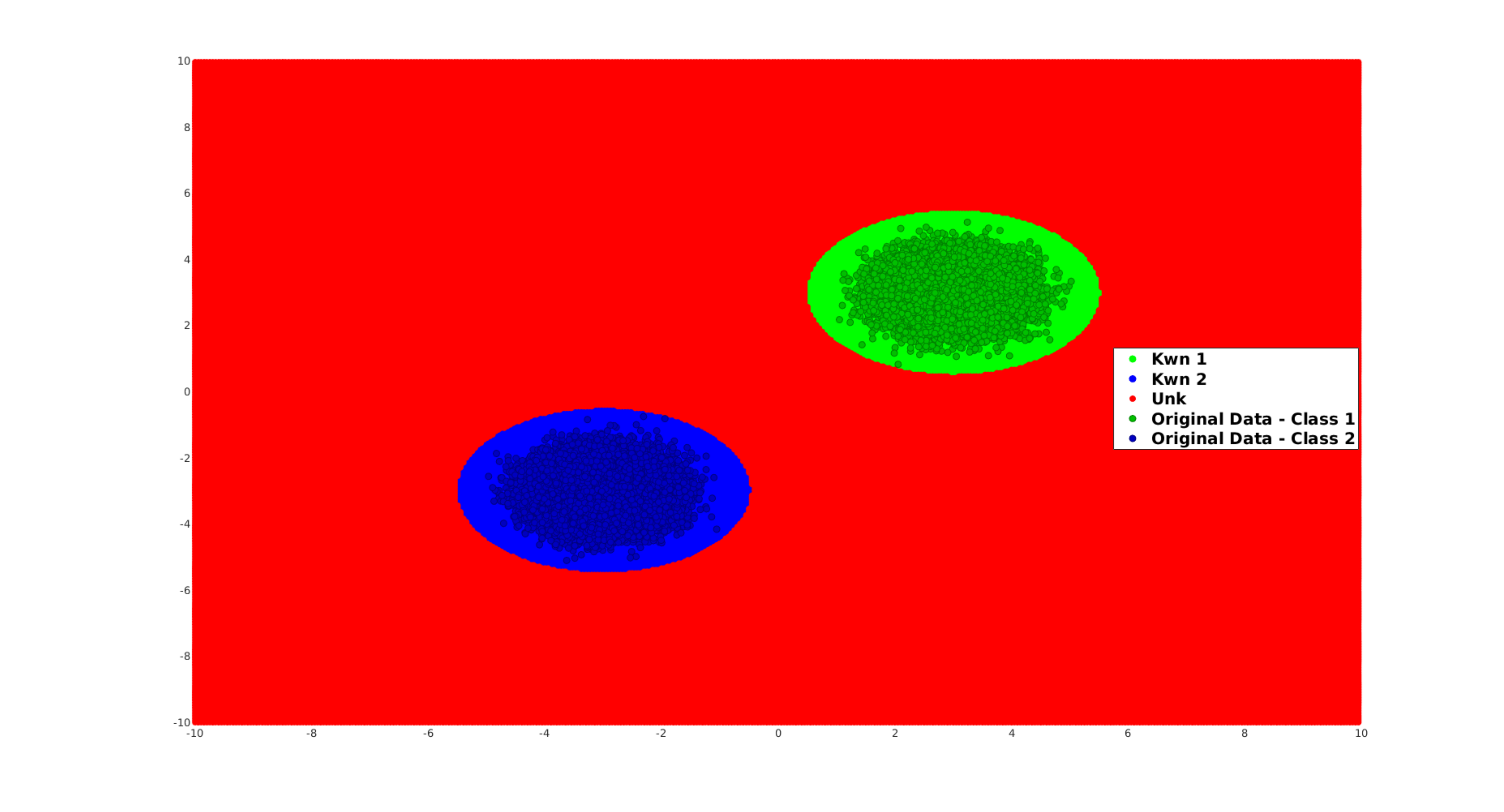}
		\vskip -0.0pt \caption{Two-Gauss}
		\label{fig:toy_two_gauss}
	\end{subfigure}
	~
	\begin{subfigure}[t]{0.32\textwidth}
		\centering
		\includegraphics[width=1.0\linewidth]{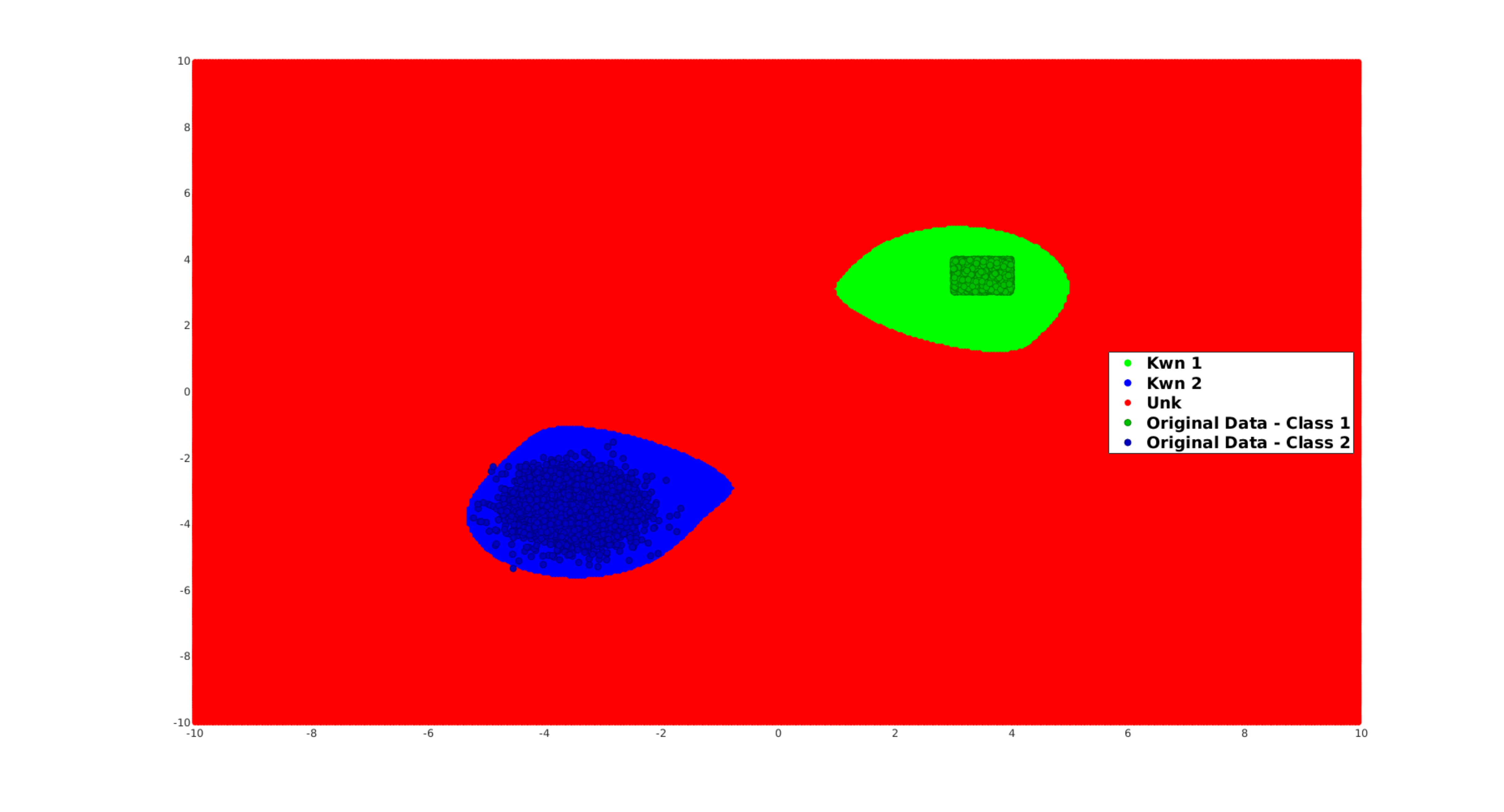}
		\vskip -0.0pt \caption{Uni-Gauss}
		\label{fig:toy_four_gauss}
	\end{subfigure}    
	~
	\begin{subfigure}[t]{0.32\textwidth}
		\centering
		\includegraphics[width=1.0\linewidth]{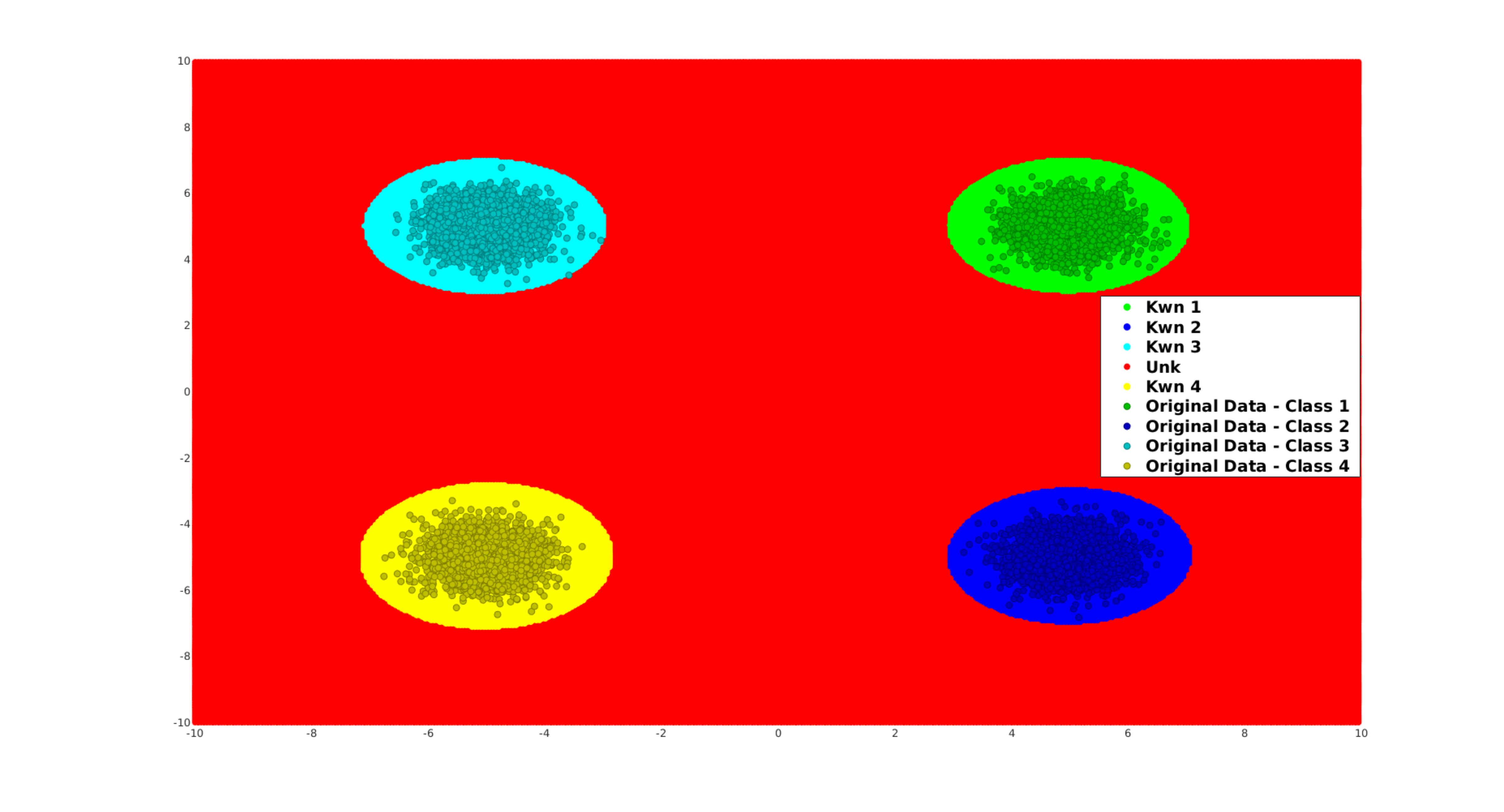}
		\vskip -0.0pt \caption{Four-Gauss}
		\label{fig:toy_multi_gauss}
	\end{subfigure}   
	
	\vskip -0.0pt \caption{Toy Examples.}
	\label{fig:toy}
\end{figure*}

To see the decision boundaries learned using the proposed approach, we perform few experiments with 2-Dimensional toy data. For these experiments the encoder, decoder and classifier architectures are FC(2)-Sig-FC(5)-Sig, FC(5)-Sig-FC(2) and FC(5)-Sig-FC(2), respectively. Here, FC(T) indicates fully connected layer with T hidden units, Sig is the sigmoid activation. We train these networks using the proposed approach for three different variations of 2-Dimensional datasets, namely Two-Gauss, Four-Gauss and Uni-Gauss. Two-Gauss and Four-Gauss have two and four 2D Gaussians with different means and same variance, respectively. Whereas Uni-Gauss has one class as 2D Gaussian and another classes Uniformly distributed. As it can be seen from Fig.~\ref{fig:toy}, the proposed approach is able to learn tight boundaries surrounding the data points and identify all of the remaining space as unknown.

\subsection{Results}
Here we present the AUROC table for open-set identification with standard deviation values. Standard deviation values were not available for CIFAR+10, CIFAR+50 and TinyImageNet as the values are taken from \cite{neal2018open}. 

\begin{table*}[!h]
	\centering
	\resizebox{\linewidth}{!}{
		\begin{tabular}{|c|c|c|c|c|c|c|}
			\hline
			\textbf{Method}   & \textbf{MNIST} & \textbf{SVHN} & \textbf{CIFAR10} & \textbf{CIFAR+10} & \textbf{CIFAR+50} & \textbf{TinyImageNet} \\ \hline
			SoftMax            & 0.978 $\pm$ 0.002 & 0.886 $\pm$ 0.006 & 0.677 $\pm$ 0.032 & 0.816 $\pm$ -- & 0.805 $\pm$ -- & 0.577 $\pm$ -- \\ \hline
			OpenMax (CVPR'16)   & 0.981 $\pm$ 0.002 & 0.894 $\pm$ 0.008 & 0.695 $\pm$ 0.032 & 0.817 $\pm$ -- & 0.796 $\pm$ -- & 0.576 $\pm$ -- \\ \hline
			G-OpenMax (BMVC'17) & 0.984 $\pm$ 0.001 & 0.896 $\pm$ 0.006 & 0.675 $\pm$ 0.035 & 0.827 $\pm$ -- & 0.819 $\pm$ -- & 0.580 $\pm$ -- \\ \hline
			OSRCI (ECCV'18)     & 0.988 $\pm$ 0.001 & 0.910 $\pm$ 0.006 & 0.699 $\pm$ 0.029 & 0.838 $\pm$ -- & 0.827 $\pm$ -- & 0.586 $\pm$ -- \\ \hline
			Proposed Method              & \textbf{0.989 $\pm$ 0.002} & \textbf{0.922 $\pm$ 0.009} & \textbf{0.895 $\pm$ 0.008} & \textbf{0.955 $\pm$ 0.006} & \textbf{0.937 $\pm$ 0.004} & \textbf{0.748 $\pm$ 0.005} \\ \hline
		\end{tabular}
	}
	\vskip -0.0pt \caption{AUROC for open-set identification, values other than the proposed method are taken from \cite{neal2018open}. Standard deviation value for state of the art not available for CIFAR+10, CIFAR+50 and TinyImageNet.}
	\label{table:auroc}
\end{table*}

\subsection{Histogram Progression}

\begin{figure*}[h!]
	\centering
	\begin{subfigure}[t]{0.33\textwidth}
		\centering
		\includegraphics[width=1.0\linewidth]{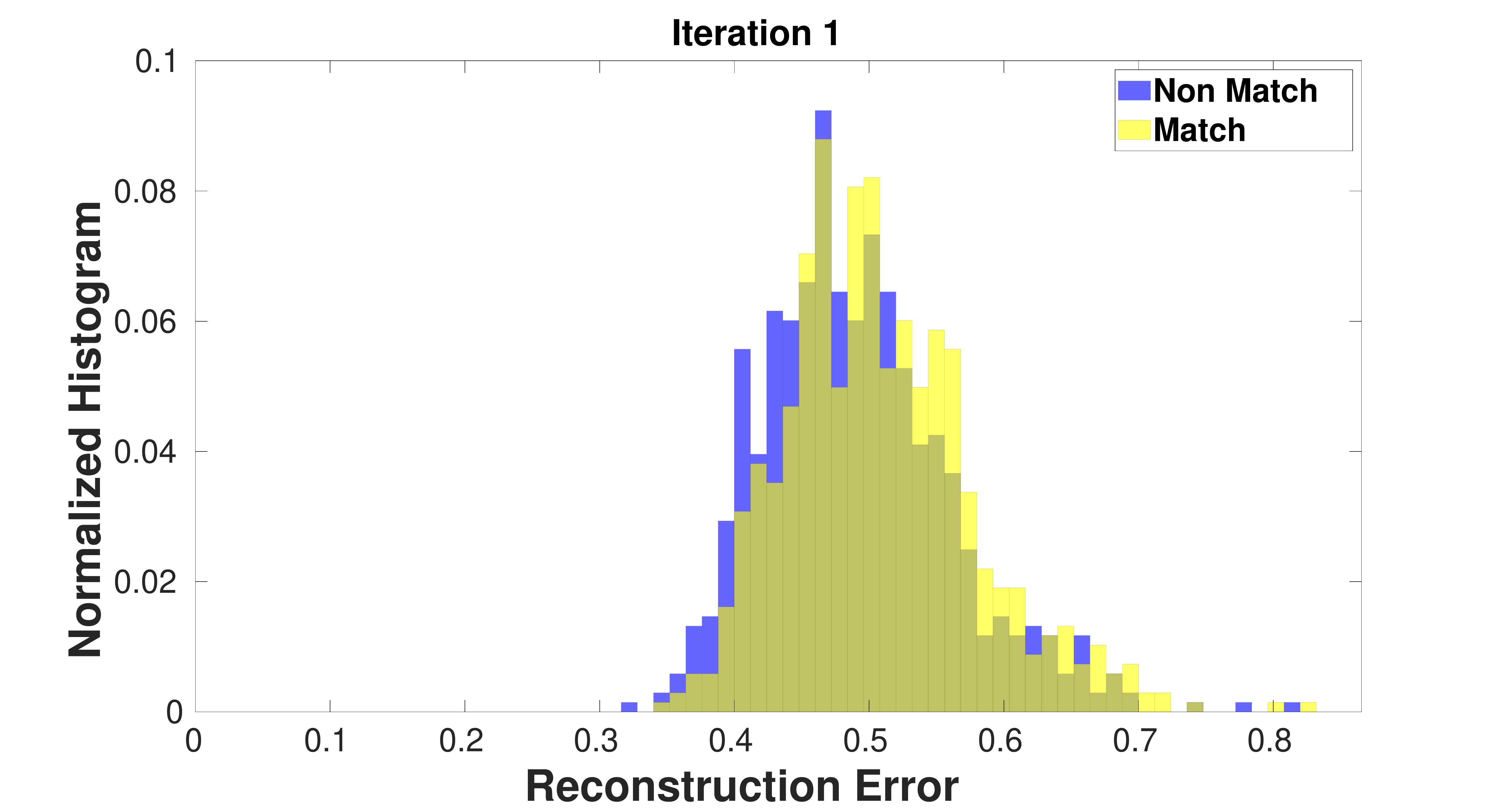}
		\vskip -0.0pt \caption{ }
		\label{fig:nm_1}
	\end{subfigure}
	~
	\begin{subfigure}[t]{0.32\textwidth}
		\centering
		\includegraphics[width=1.0\linewidth]{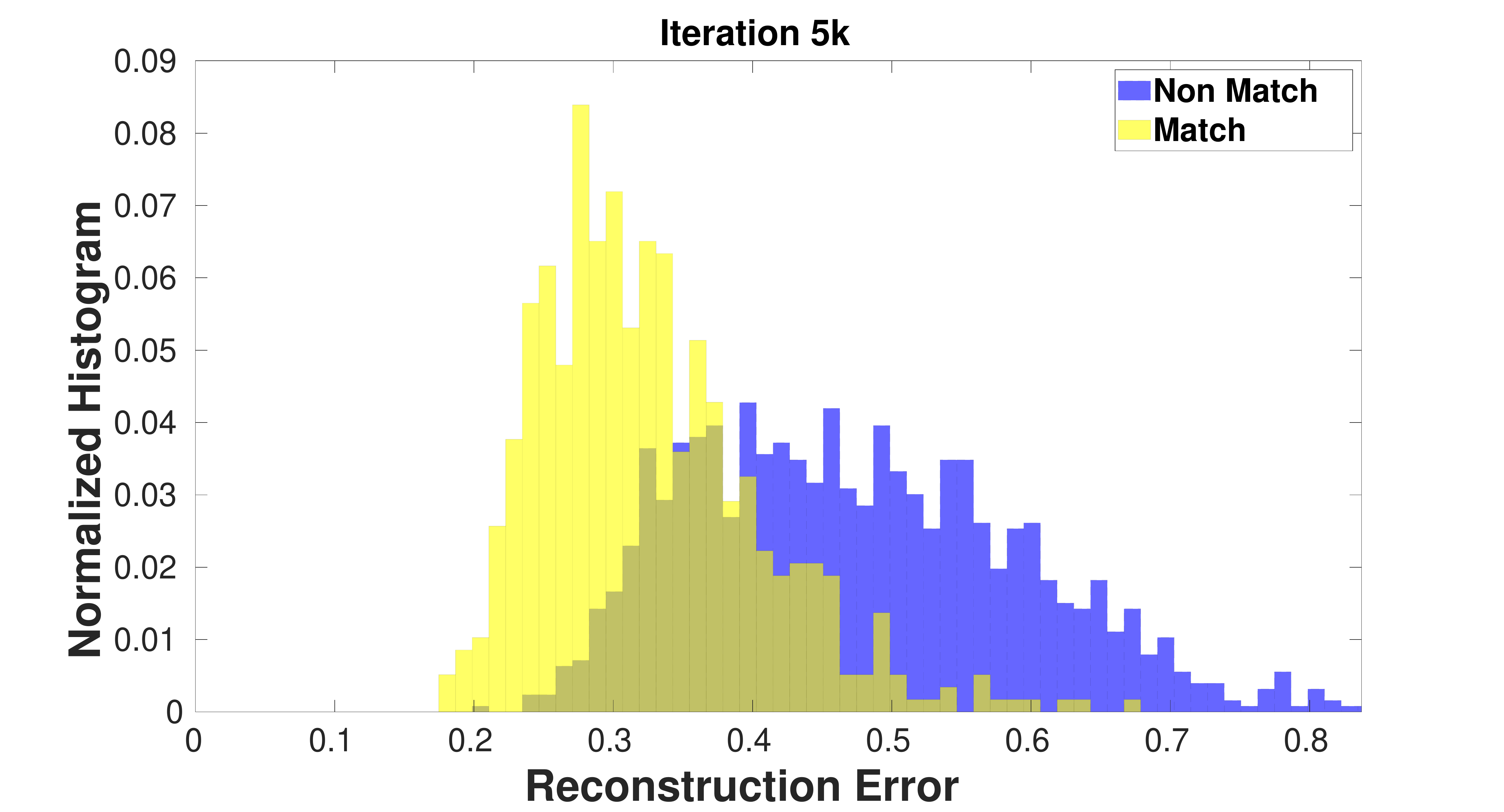}
		\vskip -0.0pt \caption{ }
		\label{fig:nm_5k}
	\end{subfigure}    
	~
	\begin{subfigure}[t]{0.32\textwidth}
		\centering
		\includegraphics[width=1.0\linewidth]{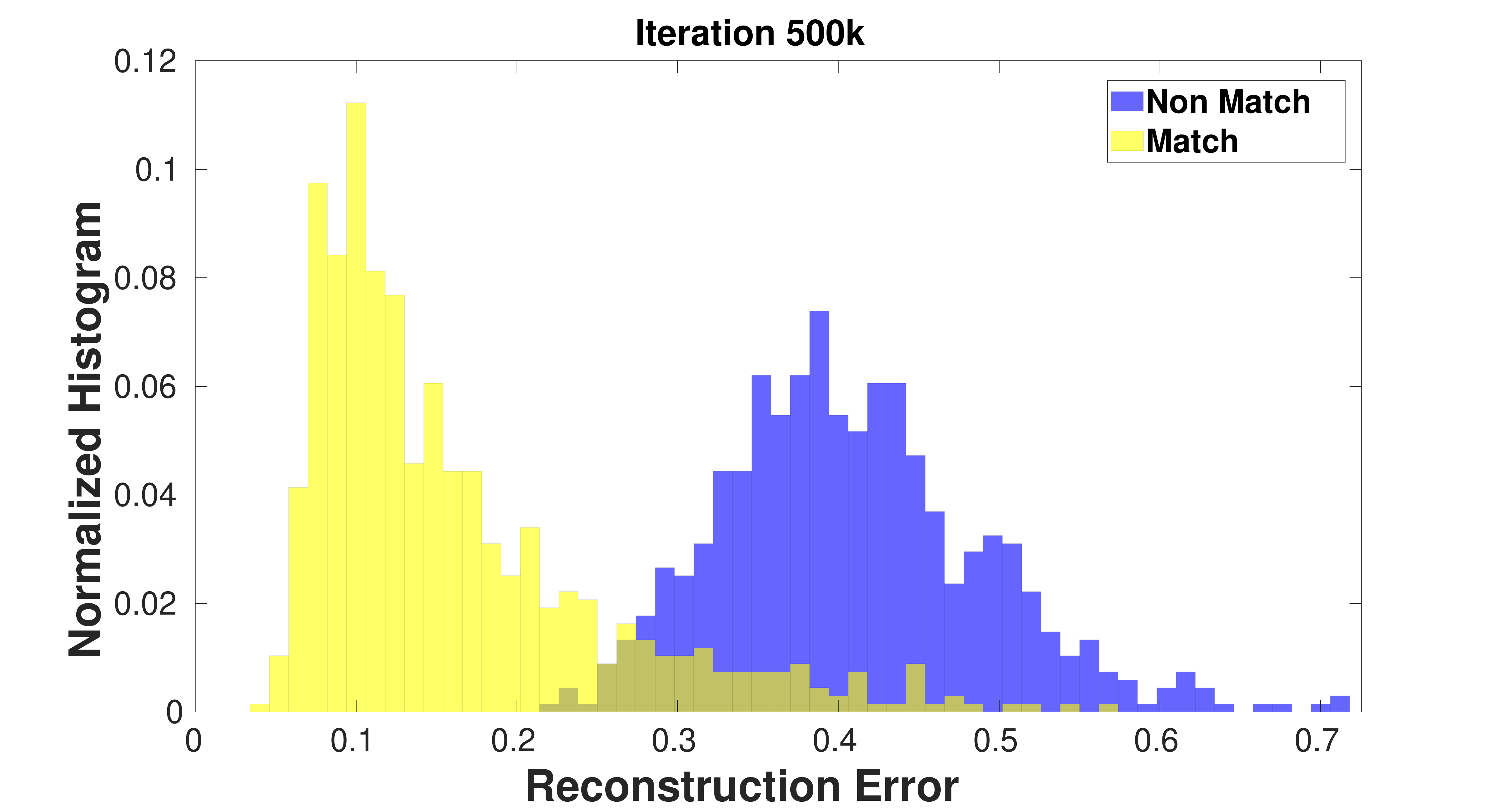}
		\vskip -0.0pt \caption{ }
		\label{fig:nm_500k}
	\end{subfigure}   
	
	\vskip -0.0pt \caption{Progression of \textbf{Match} and \textbf{Non match} data reconstruction error distribution with training iterations for SVHN.}
	\label{fig:nm}
\end{figure*}

\begin{figure*}[h!]
	\centering
	\begin{subfigure}[t]{0.33\textwidth}
		\centering
		\includegraphics[width=1.0\linewidth]{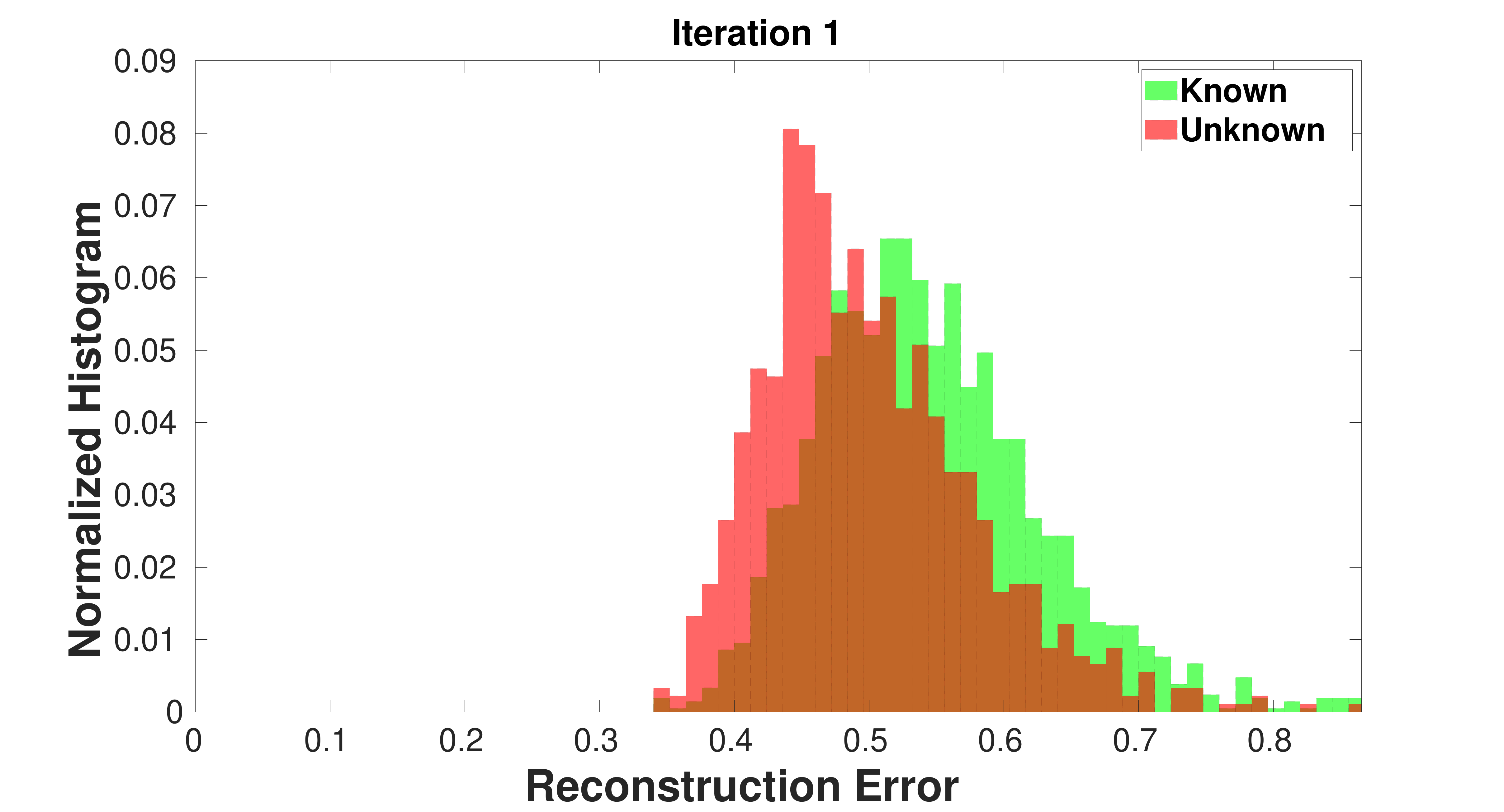}
		\vskip -0.0pt \caption{ }
		\label{fig:uk_1}
	\end{subfigure}
	~
	\begin{subfigure}[t]{0.32\textwidth}
		\centering
		\includegraphics[width=1.0\linewidth]{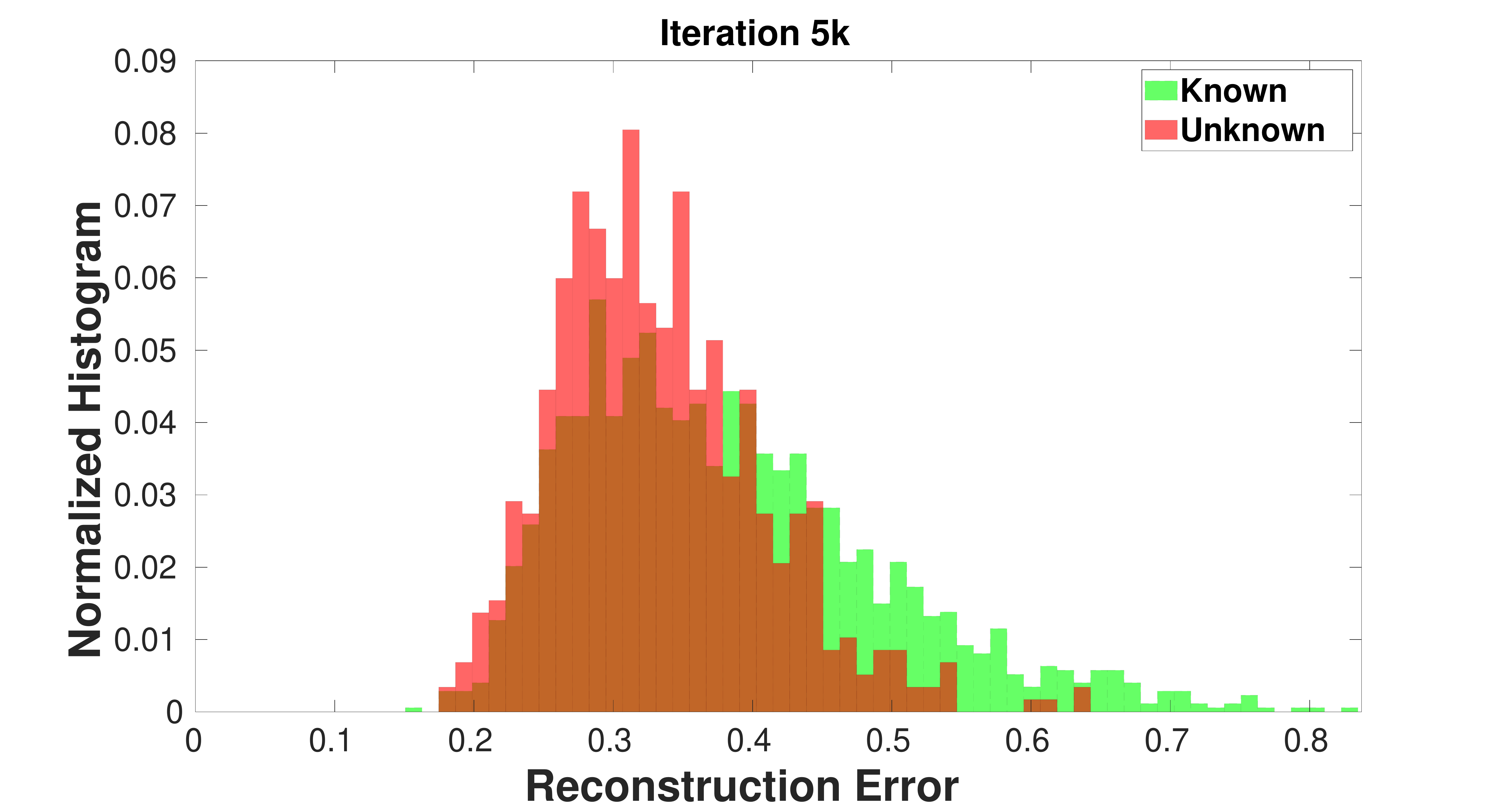}
		\vskip -0.0pt \caption{ }
		\label{fig:uk_5k}
	\end{subfigure}    
	~
	\begin{subfigure}[t]{0.32\textwidth}
		\centering
		\includegraphics[width=1.0\linewidth]{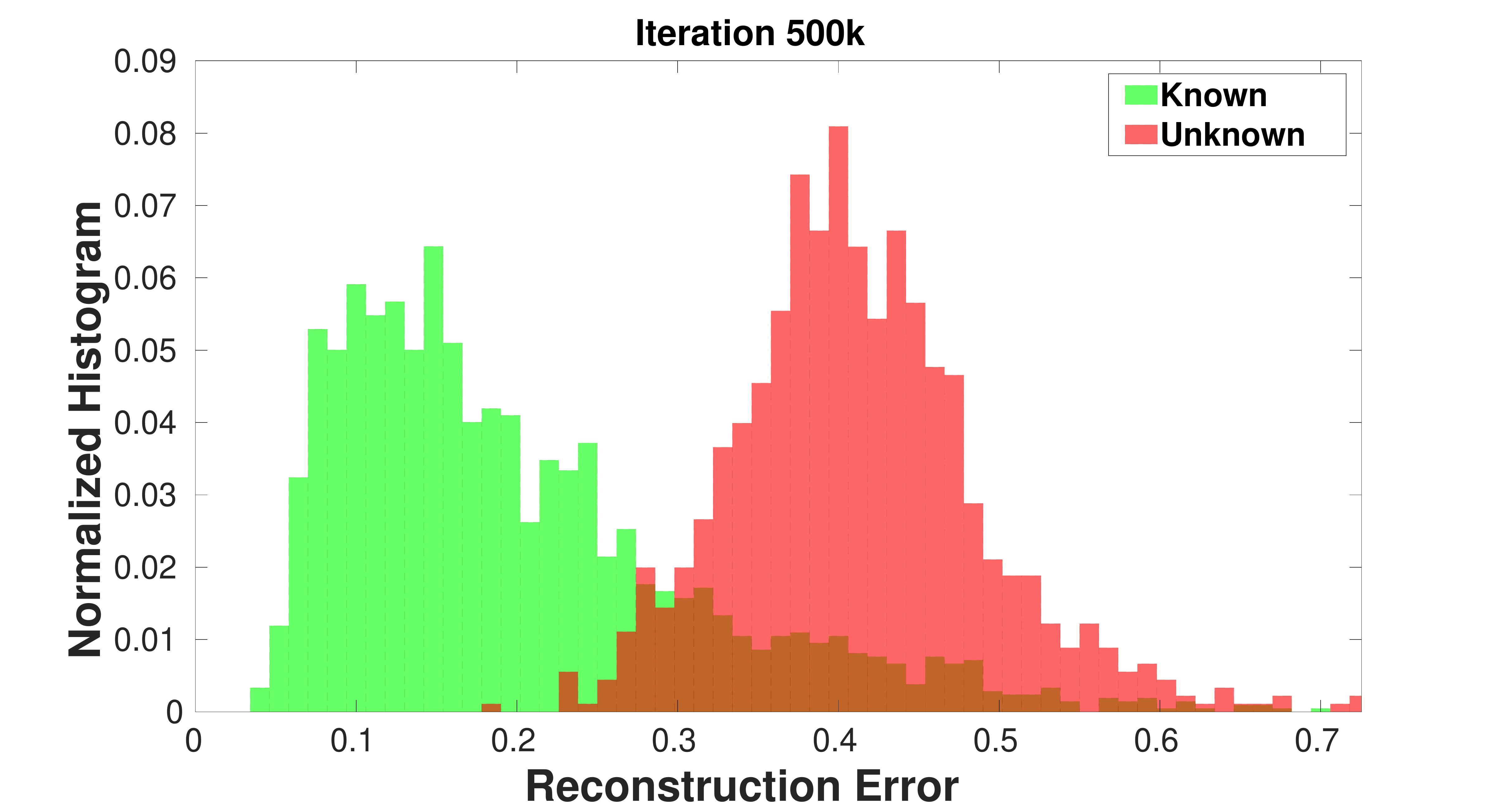}
		\vskip -0.0pt \caption{ }
		\label{fig:uk_500k}
	\end{subfigure}   
	
	\vskip -0.0pt \caption{Progression of \textbf{Known} and \textbf{Unknown} data reconstruction error distribution with training iterations for SVHN.}
	\label{fig:uk}
\end{figure*}

Fig.~\ref{fig:nm} and Fig.~\ref{fig:uk}, provides evolution of reconstruction errors during the learning procedure. The reconstruction errors for match, non match, known and unknown are provided at iteration 1, 5k and 500k. As it can be seen from Fig.~\ref{fig:nm_1}, since the network is initialized with random weights, the reconstruction errors for match and non match are not discriminative. However, since the  network is trained to learn the discriminative reconstructions for match and non match conditioning, with the increase in iterations, the reconstruction errors become more discriminative as seen from the Fig.~\ref{fig:nm_5k} and Fig.~\ref{fig:nm_500k}. As a result, known and unknown reconstruction errors follow the same trend as that of match and non match as evident from the Fig.~\ref{fig:uk}. The SVHN dataset is used for generating the normalized histograms of match, non match, known and unknown data reconstruction errors.

\newpage

\subsection{Network Architecture}

\begin{figure*}[b!]
	\begin{center}
		\includegraphics[width=1.00\linewidth]{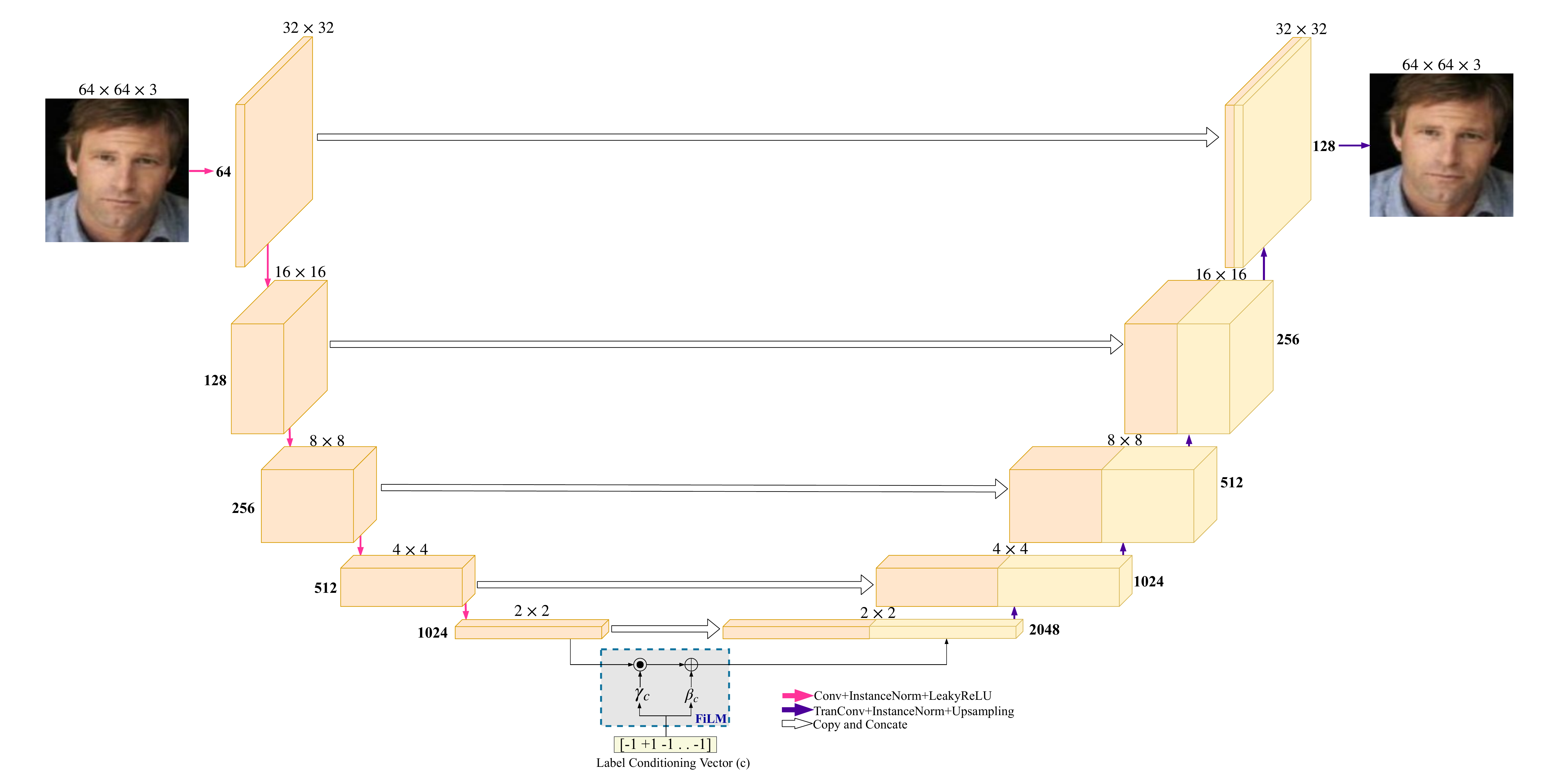}
	\end{center}
	\vskip -0.0pt \caption{U-Net based Architecture used for LFW experiments.}
	\label{fig:unet}
\end{figure*}

The network architecture for the LFW experiments is shown in the Fig.~\ref{fig:unet}. It is an U-Net inspired network architecture with a FiLM conditioning layer in the middle. The network architecture is as follows,\\
C(64)-C(128)-C(256)-C(512)-C(1024)-FiLM-DC(2048)-DC(1024)-DC(512)-DC(256)-DC(124)-DC(3)-Tanh.\\
Here, C(T) represents T channel convolution layer followed by instance normalization and leaky ReLU activation. DC(T) represents T channel transposed convolution layer followed by instance normalization and Upsampling. FiLM layer is a conditioning layer which modulates feature maps from C(1024) with linear modulation parameters $\gamma_c$ and $\beta_c$ of size 1024$\times$2$\times$2, based on label conditioning vector. Here, the convolution blocks are used as encoder and deconvolution blocks are used as decoder. As explained in the proposed approach, the encoder weights are frozen during training in stage-2. The classifier network for the experiments with the LFW dataset is a single layer fully connected network with 12 hidden units (same as number of known classes).

\end{document}